\documentclass[11pt]{article}
\usepackage{fullpage}

\renewcommand{\thesection}{\arabic{section}}
\renewcommand{\thesubsection}{\thesection.\arabic{subsection}}
\renewcommand{\thesubsubsection}{\thesubsection.\arabic{subsubsection}}

\usepackage{titlesec}
\titleformat{\section}[block]{\Large\bfseries}{\thesection}{1em}{}
\titleformat{\subsection}[block]{\large\bfseries}{\thesubsection}{1em}{}
\titleformat{\subsubsection}[block]{\normalsize\bfseries}{\thesubsubsection}{1em}{}

\usepackage{graphicx} 
\usepackage{natbib}
\usepackage{hyperref}
    \hypersetup{
        colorlinks=true, 
        citecolor=blue,  
        linkcolor=red,   
        urlcolor=cyan    
    }
\usepackage{color}

\usepackage{graphicx, centernot, amsmath, bbm, booktabs, multirow, array, capt-of, varwidth, natbib, amstext, tabularx}
\usepackage{enumitem}
\usepackage{subcaption}
\usepackage{cleveref}
\usepackage[utf8]{inputenc}
\usepackage{dirtytalk}
\usepackage{csquotes}
\usepackage{graphicx}
\usepackage{amsmath,amsthm,amsfonts}
\usepackage{lipsum}
\usepackage[utf8]{inputenc}
\usepackage{hyperref}
\usepackage{float}
\hypersetup{
    colorlinks=true,
    linkcolor=blue,
    filecolor=magenta,      
    urlcolor=cyan,
    pdftitle={Overleaf Example},
    pdfpagemode=FullScreen,
    }
\usepackage{tikz}
\usetikzlibrary{bayesnet}

\usepackage[many]{tcolorbox}   
\definecolor{main}{HTML}{5989cf}    
\definecolor{sub}{HTML}{cde4ff}     

\tcbset{
    sharp corners,
    colback = white,
    before skip = 0.2cm,    
    after skip = 0.5cm      
}                           

\newtcolorbox{boxA}{
    fontupper = \bf,
    boxrule = 1.5pt,
    colframe = black 
}

\newtcolorbox{boxB}{
    fontupper = \bf\color{main}, 
    boxrule = 1.5pt,
    colframe = main,
    rounded corners,
    arc = 5pt   
}

\newtcolorbox{boxC}{
    colback = sub, 
    boxrule = 0pt  
}

\newtcolorbox{boxD}{
    colback = sub, 
    colframe = main, 
    boxrule = 0pt, 
    toprule = 3pt, 
    bottomrule = 3pt 
}

\title{Bridging Prediction and Intervention Problems in Social Systems
\footnote{This working paper grew out of an initial 5-day workshop meeting on "Bridging Prediction and Intervention Problems in Social Systems" at the Banff International Research Station (BIRS), Canada. The workshop gathered an interdisciplinary community of researchers from computer science/machine learning, statistics, law and social sciences interested in policy and societal applications of prediction and interventions, with the goal of informing new methodology, and proposing new governance structures. 
Overall, we explored these topics cutting across social application domains such as education, healthcare, criminal justice, social services, and finance, as well as examining the roles of users (e.g. teachers, clinicians, caseworkers) and other impacted stakeholders within common institutional and societal contexts. 
}}
\author{Lydia T.~Liu\thanks{These authors contributed equally to this work and are listed in alphabetical order.}~, Inioluwa Deborah Raji\footnotemark[2]~, Angela Zhou\footnotemark[2]~,\\ Luke Guerdan,  Jessica Hullman, Daniel Malinsky, Bryan Wilder, Simone Zhang, \\
Hammaad Adam,  Amanda Coston, Ben Laufer,  Ezinne Nwankwo,  Michael Zanger-Tishler,\\ 
Eli Ben-Michael, Solon Barocas, Avi Feller, Marissa Gerchick, Talia Gillis, Shion Guha,\\ Daniel Ho, Lily Hu,
 Kosuke Imai, Sayash Kapoor, Joshua Loftus, Razieh Nabi,\\ 
Arvind Narayanan, Ben Recht, Juan Carlos Perdomo, Matthew Salganik, Mark Sendak,\\ Alexander Tolbert, 
Berk Ustun,  Suresh Venkatasubramanian, Angelina Wang, Ashia Wilson 
\thanks{This working paper includes a provisional author list to be be finalized in later versions.}
}

\titleclass{\subsubsection}{straight}
\titleformat{\subsubsection}%
  [hang]
  {\normalfont\large\itshape}
  {\thesubsubsection}
  {1em}
  {}
  []
  
\date{\today}

\newtheorem{example}{Example}
\newtheorem{casestudy}{Case Study}

\newtheorem{parable}{Parable}

\usepackage{endnotes}

\usepackage{amsmath, amsthm, amssymb, amsfonts}
\usepackage{algpseudocode}
\usepackage{algorithm}
\usepackage{lineno}

\algtext*{EndIf} 
\algtext*{EndWhile} 
\algtext*{EndFor} 
\algtext*{EndFunction} 
\algtext*{EndProcedure} 

\usepackage{graphicx}
\usepackage{bm}
\newcommand\E{\mathbb{E}}

\usepackage{xcolor}       

\usepackage{wrapfig}

\newtheorem{issue}{Issue}

\usepackage{hyperref}

\usetikzlibrary{shapes,decorations,arrows,calc,arrows.meta,fit,positioning}
\tikzset{
    -Latex,auto,node distance =1 cm and 1 cm,semithick,
    state/.style ={ellipse, draw, minimum width = 0.7 cm},
    point/.style = {circle, draw, inner sep=0.04cm,fill,node contents={}},
    bidirected/.style={Latex-Latex,dashed},
    el/.style = {inner sep=2pt, align=left, sloped}
}

\begin{document} 

\maketitle

\begin{abstract}
Many automated decision systems (ADS) are designed to solve \emph{prediction problems}--- where the goal is to learn patterns from a sample of the population and apply them to individuals from the same population. In reality, these prediction systems operationalize holistic policy \emph{interventions} in deployment. Once deployed, ADS can shape impacted population outcomes through an effective policy change in how decision-makers operate, while also being defined by past and present interactions between stakeholders and the limitations of existing organizational, as well as societal, infrastructure and context. 
In this work, we consider the ways in which we must shift from a prediction-focused paradigm to an intervention-oriented paradigm when considering the impact of ADS within social systems. 
We argue this requires an expansion of the default problem setup for ADS to move beyond prediction to consider the implications of the decision-making context, and downstream outcomes.
We highlight how this perspective unifies modern statistical frameworks and other tools to study the design, implementation, and evaluation of ADS systems, and point to the research directions necessary to operationalize this paradigm shift. Using these tools, we characterize the limitations of focusing on isolated prediction tasks, and lay the foundation for a more intervention-oriented approach to developing and deploying ADS. 
\end{abstract}

\clearpage
\tableofcontents

\clearpage 

\section{Introduction}

Automated decision systems (ADS) leverage statistical patterns in historical data to standardize complex information about individuals, making predictions about such individuals to inform more effective decision-making regarding decision subjects~\citep{richardson2021defining}. These predictions can be derived from anything as simple as a regression model on fixed, discrete numeric features to supervised learning models on pixels to an unsupervised large language model trained on text tokens.

These ADS systems are currently widely deployed, impacting processes within organizations across various sectors, from criminal justice and healthcare to education, employment, and finance -- and thus critically informing  life-altering decisions in millions of lives. 

Notable applications from various domains include:

\begin{itemize}[leftmargin=0pt, label={}]
    \item \textbf{Criminal justice:} Judges must make pre-trial decisions about release conditions that balance individual freedom with public safety. 
    Over 60\% of the U.S. population lives in jurisdictions that use Risk Assessment Tools (RATs) to inform these sentencing decisions \citep{pji2019scan}. 
    Although shown in some cases to increase judge leniency on defendants \citep{anderson2019evaluation, albright2019if}, the downstream effectiveness of RATs in safely reducing  detainment rates, while minimizing recidivism and other concerns, within the justice system has been widely questioned \citep{angwin2016machine,green2020false}. 
    \item  \textbf{Healthcare}: Sepsis is a life-threatening condition and a significant portion of mortality and healthcare costs. Some ML-based early warning systems have increased subsequent treatment \citep{sendak2020real} and demonstrated generalizability to another location \citep{valan2025evaluating};  before-after comparisons of another system indicate reduced mortality \citep{boussina2024impact}. However, the widely deployed Epic Sepsis Model has demonstrated poor performance in external validations, where it did not identify 67\% of patients with sepsis though it produced alerts on 18\% of all hospitalized patients \citep{wong2021external}, introducing alert fatigue that could undermine impacts. 
    \item \textbf{Education}: Predictive tools are increasingly used in early-warning and student success management systems to identify students at risk of failure or dropout, with mixed results \citep{feathers2021racepredictor,feathers2023takeaways}. In a six-year randomized controlled trial, the data-driven advising protocol MAAPS led to a statistically significant 7 percentage point increase in graduation rates at Georgia State University \citep{rossmanMAAPSAdvisingExperiment2023}. In a K-12 setting, an observational study of Wisconsin’s Dropout Early Warning System (DEWS) suggested weak or null effects on student success \citep{perdomo2023difficult}. 
    \item \textbf{Social services:} Social services are often under-resourced; profiling tools help caseworkers triage resources to match them with those who would most benefit. Austria developed a worker profiling tool \citep{holl2018ams}, which could help direct caseworkers' triage of active labor market programs that help jobseekers find employment. Concerns arose around gender and citizenship bias in predictions \citep{achterhold2025fairness}. An earlier large-scale pilot study randomized caseworker access to a tool that suggested an optimal active labor market program based on estimated treatment effect \citep{behncke2009targeting}, but caseworkers largely ignored the recommendations. 
\end{itemize}

There exist countless more examples -- in finance, companies like Upstart use machine learning algorithms to predict a loan applicant's ability to pay; AI-driven hiring platforms such as Pymetrics are used by companies to streamline recruitment  \citep{wells2019pymetrics}, but may also perpetuate existing inequalities in hiring practices \citep{moore2023aclu}; finally, various public service tasks from child welfare triaging~\cite{saxena2021framework, chouldechova2018case}, to social benefit fraud detection~\cite{charette2018michigan} to homeless service resource allocation~\citep{triage-rice-13,rice2018linking,rahmattalabi2022learning,kube-das-19} are regularly mediated by data-derived predictions. 

Most of these systems are based on predictive risk models that are benchmarked for predictive accuracy, without accountability mechanisms for achieving beneficial outcomes in deployment. Despite their widespread use, sometimes deployed systems nonetheless fail to improve the outcomes (e.g., health, education, justice, employment opportunity) for decision subjects, even if the predictive models achieve good accuracy on their training data.

In particular, not all jurisdictions have the data, capacity, or personnel to train their own ADS or predictive risk model, so jurisdictions often must leverage publicly available ADS or procure from vendors.\footnote{\cite{ludwig2024unreasonable} introduce a notion of the ``marginal value of public funds" (MVPF) and argue that for these reasons of portability and others, the MVPF for such algorithmic support can be infinite. } However, performance can vary widely in different developed systems targeted at seemingly similar tasks, and even within supposedly similar deployment environments. Since ADS models are usually developed for a population, task or environment that differs from the deployment context,
 many practitioners cannot rely on the provided 
 performance reporting, typically from vendors, in order to anticipate the effectiveness of the ADS model in a given scenario. 

What explains this disconnect? Why do some systems help achieve gains, but not others? We argue that these mixed results reflect structural difficulties in improving decision-making processes in institutional contexts on the basis of prediction. Simply put, systems are not deployed in a vacuum. The impact of ADS depends on how it is built, evaluated, and implemented. 
As such, the models deployed may prioritize prediction as the primary goal \citep{kleinberg16guide,kleinberg2015prediction}, without accounting for the broader contexts in which they operate~\citep{liu2023reimagining}, such as discretion of ``street-level bureaucrats" who who use them \citep{alkhatib2019street}, or proxy/delayed outcomes. 

Further, when institutions focus solely on maximizing predictive accuracy without accounting for implementation contexts, systems may not only fail to improve outcomes but harm the very people they purport to help. Sophisticated models that predict accurately yet fail to deliver meaningful improvements in practice represent a fundamental challenge that demands a new approach to how we design, evaluate, and implement these tools.

In this paper, we propose an \emph{interventional} perspective on ADS as a powerful way to address these challenges. By viewing predictive models as active components of policy interventions, researchers and practitioners can focus on how these systems influence decisions and their downstream impact on outcomes. 
\textbf{By framing predictions as interventions in broader social systems, we center the consideration of the societal impacts of these tools, ensuring that they are deployed with attention to fairness, legitimacy, and the effective treatment of individuals affected by their decisions.} 

We argue that the responsible deployment of ADS requires a paradigm shift from focusing solely on prediction accuracy to a formal understanding of their effects within broader institutional and societal policy processes. In particular, the technical foundations for model design, evaluation and implementation details of ADS should account for the context, implementation and downstream effects of such tools. This shift provides the foundation to design, deploy, and evaluate algorithmic decision making in the kinds of real-world applications where they are used. 
Our specific contributions are as follows.
\begin{enumerate}[leftmargin=*]
\item   \textbf{Model Design}: We outline decision-theoretic formulations and variations of simple prediction problems, highlighting the historical use of predictions in targeting interventions where individual treatment rules may be preferred. 
\item \textbf{Evaluation Science}: We propose a unifying evaluation framework for assessing the impacts of predictions on decisions and outcomes during deployment via causal, observational and experimental estimation, going beyond traditional train-test benchmarking in predictive ADS. 
\item \textbf{Implementation Science}: We highlight key implementation context factors beyond model accuracy and performance that are often overlooked in the consideration of what makes a prediction effective in deployment. 
\end{enumerate}

\subsection{Scope}

We focus on prediction-based interventions in social systems, specifically automated decision systems used for institutional decisions that impact human well-being and opportunity.
These systems predict (latent or future) outcomes to support decision-making. 
Certainly, there are many different sociotechnical arenas where predictions impact individuals, beyond our focus on ADS in consequential social systems: such as online platforms, management of complex systems, or social media. In comparison to these other venues, for example, we focus on the introduction of ADS into existing social processes: decision-makers interact with the algorithm's recommendations and have agency to act, while decision subjects can only respond, appeal, or contest them. Typically, there are far fewer decision-makers than decision subjects, as seen in examples like teachers and students or doctors and patients. The \textit{users} of ADS products are \textit{institutional decision makers} rather than individual consumers. Often institutions adopt ADS to operationalize a broader policy change, potentially mediated or defined by a broader societal or government policy, but not necessarily so. 

The framework we present is agnostic to the specific computational modeling approach used to generate predictions, whether through classical machine learning models or generative AI approaches.
At a time when generative AI and foundation models dominate public discourse surrounding regulation and accountability, we recognize that these systems raise fundamental design challenges and that responsible deployment remains unresolved.
However, the insights from this paper apply broadly to any predictive system used in institutional decision-making contexts, and we believe these findings to be relevant regardless of changing trends to the involved prediction method.

\begin{figure*}[h!]
    \centering
    \includegraphics[width=1\textwidth]{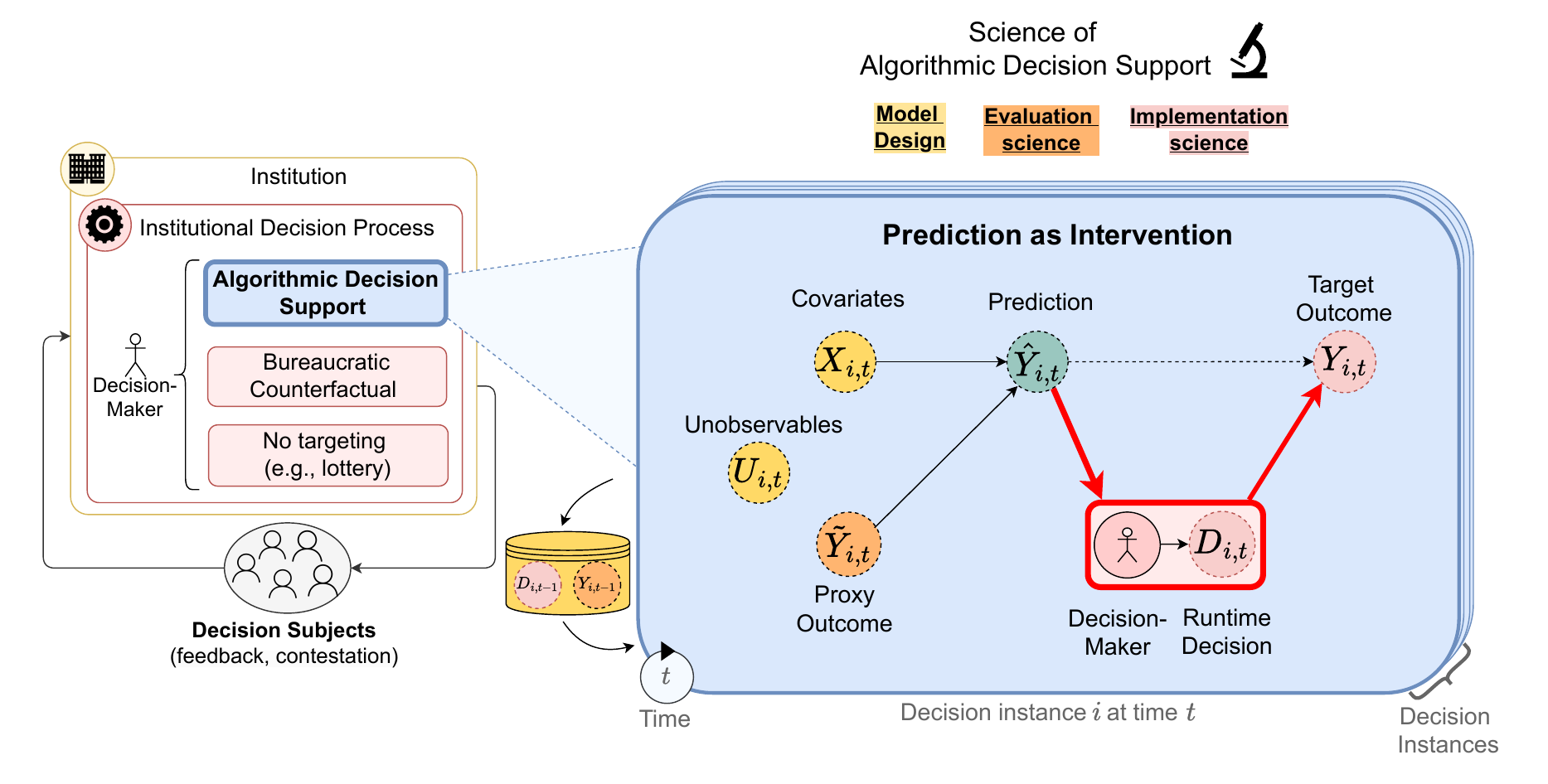}
    \caption{Conceptual diagram of the ADS universe. ADS (right hand side) are deployed in institutional contexts (left-hand side). The blue ``Prediction as Intervention" box zooms into the ADS and highlights the role of historical data on decisions and outcomes (orange database), developing new predictive models to inform future decisions mediated by human decision-makers, all in order to improve target outcomes. While most research focuses on the "prediction as intervention box", institutional context is key for understanding ADS as one alternative among others. 
    }
    \label{fig:Diagram}
\end{figure*}


\paragraph{Paper Organization} 
We start by introducing key concepts and case studies of decision support systems in \Cref{fig:Diagram}, including a discussion of relevant case studies of predictive algorithms in policy contexts. After the background section, the next three sections develop our framework of the end-to-end ecosystem of ADS: problem formulation, evaluating the impacts of ADS on downstream decisions and therefore outcomes, and the complexities of on-the-ground implementation that impact evaluation. In Section~\ref{sec:model_design}, we examine how current research addresses decision-making impacts beyond prediction, exploring various decision-theoretic approaches. In Section~\ref{sec:eval_science}, we discuss evaluation, highlighting the limitations of existing practices to evaluate ADS. In Section~\ref{sec:implement}, we survey work on deployment considerations, including intervention design, interaction design, and regulatory compliance. In Section~\ref{section-research}, we outline paths to improve the impacts of ADS by engineering, oversight, and governance. 

\section{Background}

The institutional context is crucial to understand why ADS were considered in the first place, and what are the barriers and opportunities for their beneficial impacts. We consider the institutional context a key part of the default problem setup.

\paragraph{Other surveys of ADS} Our later discussions focus on the implications of these specific aspects of ADS for their model design, evaluation and implementation science. Other works discuss ADS more generally, or with a different focus. \citet{richardson2021defining} focuses on developing definitions of algorithmic decision support for legislative and regulatory purposes. Just to name a few others, \citet{levy2021algorithms} survey the use of algorithms and decision-making in the public sector. \citep{saxena2021admaps} propose a framework (ADMAPS) focusing on sociotechnical interactions between human discretion, bureaucratic processes, and algorithmic decision-making. \citet{wang2024against} argue against the legitimacy of predictive optimization in consequential domains. 

\subsection{Details of the Institutional Context}
\label{sec-background-institutional context}

\paragraph{Goals, Resources, Activities}

Organizations deploy predictive risk models in institutional context of their goals, processes, and constraints. We consider the institutional context to be an  important part of our problem formulation. 

\begin{figure}
    \centering
    \includegraphics[width=0.75\linewidth]{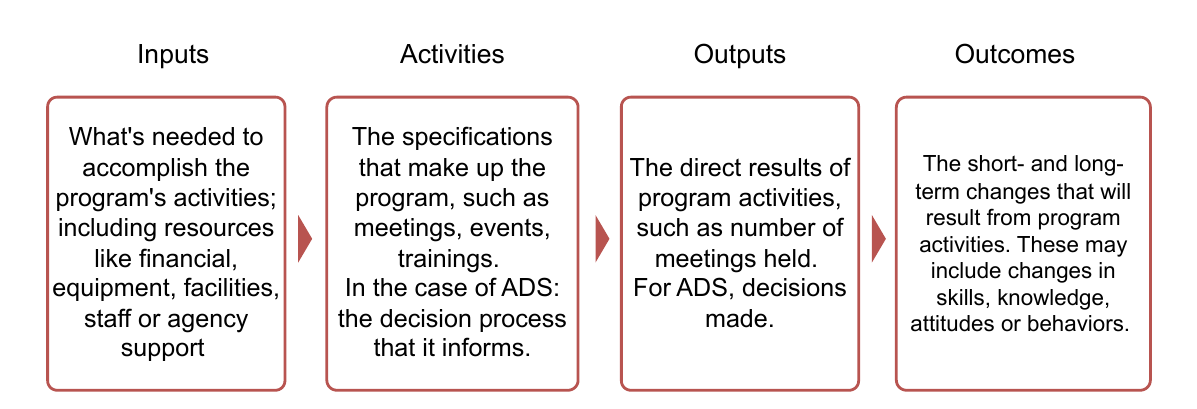}
    \caption{Program logic models (illustrative example)}
    \label{fig:program-logic-model}
\end{figure}

To illustrate what we mean by institutional context, we introduce program logic models from the program evaluation literature. Program evaluation is the field of evaluating causal impacts of interventions, social programs, or processes, which were often devised to improve social conditions. 
A program logic diagram (example in \Cref{fig:program-logic-model}) simply connects organizational resources and processes with their ultimate goals.\footnote{At the very left/beginning of the diagram, the inputs to the program are resources: such as staff (who may be overseeing recommendations), available data, or other resources. At the very right are ultimate policy goals, which a program (for example, introducing a new predictive model into a decision-making process) aims to influence and make progress towards. For example, prediction models are often used to triage limited resources; but overall policy goals might be better achieved by increasing the amount of resources rather than working on the margin to re-allocate who receives limited resources. 
In between, the organization conducts  programmatic activities: these simply comprise the interactions of organizations with individuals, including observation/measurement and other decisions. In medicine, patients receive care; in criminal justice, individuals receive pre-trial services and decisions about supervision, release, or detention; interventions; in social services, clients engage with social workers who remember their personal history and new goals, connect them to resources, advocate for and assess them, and help them navigate services. 
} It simply reminds of the broader context of inputs (resources and constraints), activities (decisions and interventions), and a wide variety of outcomes ranging from short or longer-term that might be plausibly directly affected by the deployment of predictive models (and their algorithmic systems). 


What we call the \textit{total cost of ownership} of ADS refers to the substantial resource investment required to formulate, develop, evaluate and maintain a predictive model and/or ADS. Since organizations in consequential domains are constantly tasked with doing more with less, a key question is\textit{whether the total cost of ownership of ADS is worth it, in terms of improving societal outcomes, compared to alternatives.} It is urgent to establish a careful science of evaluating ADS' impacts so that institutions can actively steer ADS development, rather than being steered by hype or inertia.

\paragraph{Policy Context}

Beyond the institutional context of a single organization and its activities, predictive models are deployed in broader policy contexts, which shape their impacts in important ways. ADS may enforce status quo policy by another hand, or lend legitimacy to and operationalize austerity policies and cost-cutting measures that maintain resource scarcity. Their introduction may be precisely \textit{because of} policy changes and aligned reform movements, making it difficult to isolate the specific effects of the predictive tools themselves.
For example, for risk assessment instruments (RAIs) in criminal justice, it is difficult to disentangle potential impacts of RAIs on reducing racial disparities from simultaneous braoder prosecutorial or bail reform movements.
Conversely, attempts to enact reforms by \textit{computational} changes to ADS are brittle: if front-line decision-makers disagree with the goals, they might simply ignore ADS information.

\begin{example}[Risk assessment for pretrial decision-making]\label{ex:justice}

When someone is charged with a criminal offense and placed in jail, judges decide whether the arrested person should be released before trial and under what conditions. Pretrial risk assessments aim to help judges make these decisions by predicting whether a defendant will fail to appear for future court dates or commit a crime if released. Although laws vary across U.S. jurisdictions, judges are typically required to weigh an individual's risk of flight and future criminal activity when choosing from several pretrial options: unconditional release, release with non-monetary conditions like supervision or drug testing, money bail (which requires defendants to pay a set amount to secure release or remain in jail), or detention without the possibility of release. These consequential decisions often must be made quickly and with limited information about the defendant. 
\end{example}

Pretrial risk assessments have been advanced as potential correctives for several long-standing issues in this challenging decision-making environment. By purporting to help judges more accurately assess a defendant's risk, risk assessments have been suggested as tools to address over-detention and over-reliance on cash bail — widely criticized for tying release to financial means — as well as racial biases in decision-making and public safety concerns. A substantial body of scholarship and public writing has critiqued risk assessments on legal, ethical, and practical grounds (for reviews, see \cite{eckhouse2018layers, robinson2019civil}), yet they remain key components of reform efforts. Most notably, when New Jersey essentially eliminated cash bail statewide in 2017, it required judges to consider an algorithmic risk assessment in their decision-making while retaining judicial discretion over final pretrial determinations.

\paragraph{Bureaucratic Counterfactuals} 
Algorithmic decision support is just one available choice, out of others, including a status quo without ADS. 
\citet{johnson2022bureaucratic} introduce the idea of a ``bureaucratic counterfactual": in the absence of a predictive model, organizations would use manual rule-based methods based on prioritization of categorical eligibility. 

For example, in pretrial risk assessment, the ``bureaucratic counterfactual" without any ADS comprises of judges' own discretion. An important question is whether decisions with ADS may be more or less biased than decisions without. 
In housing, \citet{johnson2022bureaucratic} study prioritization for housing vouchers and find that priority categories reflect a loose mix of values such as enforcing traditional ``deserving poor" distinctions, prioritizing local community ties, and addressing health needs, reflecting multiple, sometimes competing goals for assistance rather than a formally specified basis of allocation around specific housing outcomes.

Therefore, current decisions made in these organizations reflect a combination of institutional policy, categorical prioritization, and human discretion. In  \Cref{fig:Diagram}, this is represented via the database of prior decisions and outcomes ($D$,$Y$). 

\paragraph{Institutional context as ADS deployment context. 
}

Institutional context and their ongoing practices introduce idiosyncratic variability in the requirements and goals of shared predictive tasks like ``predict heart disease" or ``predict recidivism". A general task like ``predict heart disease" becomes ``predict heart disease when doctors and hospitals have protocol A,B,C for tests and medications, and the patient case mix is D." We highlight case studies that illustrate how ADS differs from other prediction tasks.  

\begin{example}[Medical cost prediction]\label{ex:medical_cost}
In many healthcare systems, high-risk care management programs are used to allocate additional resources to patients with the greatest medical need. Given limited capacity in such programs, healthcare providers and insurers often rely on algorithmic tools to prioritize enrollment \cite[e.g.][]{Optum}. Widely deployed tools predict patients’ future healthcare \emph{costs} as a proxy for their underlying \emph{health status}, as costs are readily available in billing records and assumed to correlate with need~\citep{obermeyer2019dissecting}.

\end{example}

Institutional context dictates what data is collected in the first place---usually primarily for administrative purposes, rather than downstream research.\footnote{For example, in \Cref{ex:medical_cost} (cost prediction), a machine learning researcher might consider cost prediction rather than outcome prediction, as in \citep{obermeyer2019dissecting}, to be an error of naivete rather than a natural consequence of how institutions generate data. A lot of data collection in healthcare is to support insurance billing, and was not collected in the first place to support downstream predictive innovation or future machine-learning papers. }

\begin{example}[Triaging and targeting social services, in housing: VI-SPDAT] \label{ex:housing}
    Many jurisdictions face the impossible challenge of deciding how to allocate severely resource-limited housing supports among many vulnerable individuals in need. The vulnerability index-service prioritization decision assistance tool (VI-SPDAT) was developed in 2013 by OrgCode to assess acuity/need and has been widely used as a screening and triage tool. 
    Standardized assessment of vulnerability and need can improve upon alternative allocations based on first-come-first-served, luck, or potentially unevenly distributed caseworker discretion.
    Many cities opted to use VI-SPDAT in response to the U.S. Department of Housing and Urban Development (HUD) in 2013's requirement to standardize assessment practices and ensure that the most vulnerable people of greatest need are prioritized for receive housing. 
    However, the tool never received additional resources to support its deployment at scale. Evaluations of VI-SPDAT indicate that these tools may not capture the full level of vulnerability for particular individuals; recent evaluations show worse outcomes for certain subpopulations \citep{petry2021associations} leading to OrgCode’s announcement that the tool should be phased out of use. 

\end{example}

   \Cref{ex:housing}, VI-SPDAT for housing allocation, illustrates how ADS are embedded in casework, service provision, street-level bureaucracy, and other high-touch interactions in between a decision-maker and a decision subject. Though discussions of ADS often conjure algorithmic efficiency, the VI-SPDAT assessment is a lengthy hourlong conversation between a caseworker and a client with many sensitive questions about stigmatized risk factors or personal history.

\section{Model Design}\label{sec:model_design}

We begin by examining different problem formulations for algorithmic decision-making, contrasting approaches that rely on \emph{predicting risk} alone with those that aim to \emph{target interventions} directly.  To support this comparison, we formally introduce our problem setup within a causal inference framework, enabling us to study the impacts of decisions and interventions (e.g. access to additional resources in Example~\ref{ex:housing} or preventive care in Example~\ref{ex:medical_cost}) on downstream outcomes. We discuss heterogeneous treatment effects that might motivate targeting interventions on expected efficacy, rather than predictive risk. 

To ground this discussion, we present case studies that illustrate the tradeoff between using prognostic models to prioritize individuals and directly optimizing for interventional effectiveness. Assessing the sufficiency of pure prediction rather than a decision-theoretic formulation hinges on the structure of an institution's utility function---specifically, whether there is enough domain knowledge to impute the utility or cost under alternative decisions. 

\begin{boxD}
\paragraph{Key Takeaways} 
We argue that in the ADS model design process, it is necessary to map out how predictions inform decisions and outcomes in order to choose among several problem formulations: pure prediction, decision-aware learning, or optimal individual treatment regimes.  We also recognize different factors that affect problem formulation beyond purely technical considerations. Finally, even if pure prediction is sufficient, the interventional framework can be used to reason about data issues that can improve estimation of prediction models. 
\end{boxD}

\subsection{Setup: Predictions and Interventions}

Although predictive risk models have a simple formulation---predict $Y$ from $X$---they in fact are used in a wider variety of ways to impact and change future real-world outcomes. 

ADS directs additional resources, scrutiny, services, protection/prevention, or sanction. These decisions often use individual-level information to make targeted decisions and need not be data-driven. These distinct use cases of ADS often \textbf{imply different appropriate technical formulations beyond pure prediction}. Later we discuss such prediction-intervention mismatches that arise in criminal justice (supervised release \Cref{ex:supervised-release}) and housing (Example~\ref{ex:housing}).

Therefore, our default problem setup includes both predictions and the decisions that they inform in practice. We introduce the following notation: 
\begin{itemize}
    \item $X \in \mathcal{X}$: collected covariates or contextual information about individuals, where $\mathcal{X}$ is the space of all possible $X$. 
        \item $D \in \mathcal{D}$: decision(s) made by institutions about individuals, where $\mathcal{D}$ is the set of allowable actions. 
    \item $Y$: outcomes of interest, potentially affected by decisions. 
    \end{itemize}
Typical data focuses on $X$ and $Y$ alone (in the implicit absence of decision $D$) from which one can obtain predictive scores, risk models, and classifiers:
    \begin{itemize}
    \item Let $R(x)$ denote the true risk of an outcome for an individual with covariates $X=x$. 
    In particular, when $Y$ is binary, we refer to probabilistic prediction models $R(X) = P(Y=1\mid X)$; else when $Y$ is continuous, we refer to regression functions $R(X) = E[Y\mid X].$ We denote the predicted risk by $\hat R(x)$.

    \item For an individual with covariates $X = x$, and given a threshold $t$, a binary classifier can be defined as $\hat Y = \mathbb{I}[\hat R(x) > t]$, $\hat Y \in \{0,1\}$, 
    with natural extensions to multi-class or other structured outputs via analogous thresholding rules.

\end{itemize}

\paragraph{Interventional framework}
Many discussions about the impacts of algorithms implicitly invoke the presence of decisions. Sometimes as with binary classifications, decisions are direct algorithmic consequences of predictions. But in practice, decisions reflect a blend of human judgment and predictive information.

We introduce causal inference to explicitly model decisions 
as treatment assignments whose effects are uncertain. Each subject is described by pre-decision covariates $X$, realized decision(s) $D$, and realized outcome(s) $Y$. Under the Neyman-Rubin potential outcomes framework \citep{rubin2005causal}, we posit a vector of potential outcomes $Y(d)$, $d \in \mathcal{D}$, where $Y(d)$ denotes the outcome that would obtain under intervention $D=d$. 
In practice, we only observe the realized outcome $Y = Y(D)$, giving rise to the
``fundamental problem of causal inference." 
For the rest of this section, we assume for brevity that the decisions are binary, i.e. $d\in\{0,1\}$, though the framework extends naturally to multi-valued decision variables.

A common estimand in causal inference is 
the average treatment effect (ATE), defined as: 
$$ \textrm{ATE} := \E[Y(1)-Y(0)]. $$
The average treatment effect quantifies the impact of treating everyone $\E[Y(1)]$ v.s. treating no-one $\E[Y(0)].$
We can define analogous risk scores and outcome models that predict potential outcomes from covariates. The conditional average treatment effect (CATE), a measure of heterogeneous treatment effect, is a covariate-conditional expected difference in outcomes under treatment ($D=1$) versus control ($D=0$): 
$$
\textrm{CATE}(X) := \E[Y(1)-Y(0) \mid X]. 
$$

Conclusions from causal inference are only as reliable as their underlying \emph{assumptions}. Since outcomes for decisions not previously made are not observed in the data, they must be imputed via statistical estimation and causal assumptions. The task of \textbf{causal identification} is that of writing down a target estimand (such as the ATE or CATE) in terms of probability distributions on observables. Estimation, which we do not discuss in detail, is a separate task after identification. We briefly describe the typical assumptions for causal identification and estimation, in our particular context of decision-making:

\begin{itemize}
  \item \textit{Consistency and no interference.}  \textit{Consistency} states that the outcome observed under a particular decision corresponds to the potential outcome under that decision—formally, $Y_i = Y_i(d)$ if $D_i = d$. A related assumption is \textit{no interference}, meaning that an individual's potential outcomes are not affected by the decisions assigned to others, $D_j, j\ne i$. These assumptions are often jointly referred to as the \textit{Stable Unit Treatment Value Assumption (SUTVA)}.

  \item \textit{Overlap (\textit{positivity}).}  This assumption requires that the probability of receiving either treatment or control, given covariates, is strictly positive: 
  \[
  P(D = d \mid X=x) \geq \nu,\quad \text{for } \ \nu > 0, \ d \in \{0, 1\}, \ \text{and all } x \ \text{where } P(X=x) > 0.
  \]
  This condition ensures that for every combination of covariates, both decisions are possible. If there is no chance of observing a particular decision for an individual, then there is no information available to estimate the decision’s impact for that individual.

  \item \textit{Unconfoundedness (\textit{ignorability}).} The assumption posits that
  \[
  Y(d) \perp D \mid X, \quad d \in \{0, 1\}. 
  \]
  That is, conditional on observed covariates, decisions are “as-if” randomized. Under this assumption, differences in outcomes can be attributed to differences in decisions after adjusting for covariates, rather than to unobserved confounders that influence both treatment and outcome. While this assumption holds by design in randomized trials, it is generally only approximated in observational studies. It may fail, for instance, if treatment decisions were informed by unmeasured factors, such as a human decision maker's assessment (e.g. judge, frontline care worker) of the subject’s demeanor.
\end{itemize}

The ATE is itself the expectation of a conditional-average heterogeneous treatment effect (CATE). That is, by iterated expectations and these causal assumptions (consistency, no-interference, and unconfoundedness), we can identify the ATE as an average of CATE, which is in-turn identified as the difference between conditional outcome models.\footnote{Identification as an average of outcome models is often termed \textit{regression adjustment}, and the last formula here can be evaluated via averaging the difference of regression models over the covariate data.
} 
$$\text{ATE} 
= \E[\E[ Y(1) - Y(0)\mid X]] = \E[\E[Y|D=1,X]-\E[Y\mid D=0,X]].$$

We emphasize that it is exactly these (sometimes untestable) causal assumptions that enable causal identification, relating the causal potential outcomes to what is observed from data. This poses challenges for studying ADS as these assumptions are often violated, requiring sensitivity analysis. We gloss over many other details on identification and estimation. See various tutorials and books for more detail \citep{imbens2015causal,hernan2010causal,kennedy2016semiparametric}.

In any given context, one part of this framework---such as pure prediction of $Y$ with $\hat{R}$ or purely interventional problems based on CATE---may be more relevant than others. Overall, we find it useful for highlighting alternative problem formulations.

\paragraph{Predictions to decisions}
While we use $D$ to represent historical decisions, we consider alternative decision-making procedures, summarized in the \textbf{decision policy} $\pi(X) : \mathcal{X}\mapsto \mathcal{D}$. 
Predictive risk modeling often triages scarce resources. When used as such, often the decisions are thresholded based on predictive risk, 
\begin{equation*} \pi(X) = \mathbb{I}[R(X)>t], \tag{threshold decision policy}\end{equation*} for some threshold $t$. The decision threshold can arise from an organization's budget/resource constraints, or from trading off asymmetric false-positive/false-negative costs in a common binary classification formulation of the problem. Thresholding predictive risk scores is a pervasive way of informing decisions with predictions, but not the only problem formulation. 

Predictive risk models may be learned on data in the absence of any intervention ($D_i=0~ \forall i$). When they are used to prioritize resources, often this is called \textit{targeting based on the prognostic (baseline) risk}. 
In a purely interventional framework, optimal decisions threshold on the causal effect of the intervention. In an individualized treatment regime, also called \textit{optimal treatment policy}, the optimal treatment policy maximizes the average outcomes after decisions $E[Y(\pi)]$ (when outcomes are positive benefits). Without any functional form restrictions, the optimal treatment policy thresholds on the heterogeneous causal effect, in an unbudgeted setting. If instead there were a budget constraint on how many can be treated, the optimal decision rule is a threshold on CATE (e.g. a $1-b\%$ quantile of CATE if there is a $b\%$ of population budget constraint). We contrast these two targeting policies below.\footnote{\citet{haushofer2022targeting} contrasts these in the application of anti-poverty program design, referring to the policies as “targeting on deprivation” ($\pi_{\text{risk}}$) versus “targeting on impact.” ($\pi_{\text{ITR}}$)} 
\begin{align*}
\pi_{\text{risk}}(X)&=\mathbb{I}[\E[Y(0)|X]>t]. \tag{targeting on baseline risk}\\
\pi_{\text{ITR}}(X) &= \mathbb{I}[ \E[Y(1)-Y(0)|X] >t] \label{eq:itr}\tag{optimal treatment policy}
\end{align*}
The interventional framework makes clear that prognostic targeting only considers one half of the story. These types of optimal \emph{individualized treatment rules} (ITRs) were proposed early on in the policy setting by \citep{manski2004statistical}, although recent years have seen rapid developments in instrumentation, development \citep{athey2021policy,zhao2012estimating,kallus2021minimax} and/or deployment of these treatment allocation rules, in e-commerce \citep{dudik2011doubly}, public-sector or healthcare settings.


In between prognostic predictive risk modeling and ITRs are a class of models referred to as \emph{counterfactual risk prediction models} \citep{coston2020counterfactual,keogh2024prediction}. These predict outcomes under counterfactual scenarios corresponding to hypothetical interventions. If there are two intervention options under consideration ($D=1$ or $D=0$), there may be two risk models: expected outcomes under each intervention, $\E[Y(1)\mid X], \E[Y(0)\mid X].$ 

In practice, we find that purely-predictive formulations that target on $\E[Y\mid X]$ are more common, i.e. learning a predictive model based on all prior data without any explicit modeling of prior decisions and interventions. However, if prior data does includes other decisions or interventions, this estimand hides the resulting heterogeneity, and is vulnerable to distribution shifts in interventions (including those across jurisdictions when an ADS developed in one is used in another). For example, a prior pilot study may not continue or might in fact scale up. Learning $\E[Y\mid X]$ from the pooled historical data would be brittle to such shifts, while the counterfactual risk models discussed above would remain robust\footnote{\citet{koepke2018danger} argue that ignoring recent reforms or other changes in the social world could lead to ``zombie predictions" that overestimate risk. These interventions in the historical data include pretrial services that reduce rearrest risk or bail reforms that reduce pretrial detention and iatrogenic rearrests. }.

\subsection{Predictive vs. Interventional targeting}

We have outlined potential approaches for targeting decisions based on prognostic risk $\E[Y(0)\mid X]$ (in the absence of intervention), on interventional effects $\E[Y(1)-Y(0)\mid X]$, self-selection or no targeting, and predictive risk $\E[Y\mid X]$ (marginalizing over any historical interventions). How should ADS deployers choose among them? We highlight different tradeoffs in terms of the objectives for each strategy is optimal, decision utility, ease of estimation, and complexity of deployment.

Intervention-aware targeting offers many different options beyond predictive risk models of $\E[Y\mid X]$ alone. Recent discussions of predictions vs. interventions include \citet{barabas2018interventions}, which argues that the use of predictive risk models is a poor fit for improving social conditions, drawing on the case of risk assessments in criminal justice.\footnote{ \cite{barabas2018interventions} argues that predictive models that focus on merely predicting future outcomes neither manage long-term goals of reducing reoffense (especially when interactions with the criminal justice system are iatrogenic, increasing criminal activity), nor reduce criminal risk, nor improve understanding of root causes of criminal behavior that could be used to design interventions to reduce risk. This critique, among others, points out the urgency of going beyond prediction to consider alternative avenues for improving social outcomes. } Prediction by itself doesn't directly achieve long-term goals or inform interventions to reduce risk, as argued in \citet{liu2023actionability}. One can also reconceptualize predictive risk models in terms of the subjects of prediction.\footnote{\citet{meyer2022flipping} highlight how current risk prediction focuses narrowly on harms individuals may pose to systems (e.g., future crime); by “flipping the script,” they show how prediction could just as well be used to predict system harms to individuals, such as lengthy sentences. So much more is possible beyond current formulations.}

There can also be good reasons to target based on baseline risk $\E[Y(0)\mid X]$ alone. Prioritizing the neediest, riskiest, or most vulnerable (in the absence of treatment) may align with intuitive notions of need or desert. But it may introduce tradeoffs relative to the utility gains from interventional targeting. Existing processes may align with unstated social objectives, such as reserving resources for the most vulnerable or for those who most benefit, that had not been explicitly deliberated or discussed. But these should be made explicit for deliberation.

 
An example of targeting interventions with baseline predictive models includes the use of the PSA\footnote{For example, a decision matrix on the Public Safety Assessment risk scores has been widely circulating, i.e. available on the official website and included in jurisdictional documentation of pretrial release systems. However, it is unclear how these thresholds were determined.} for supervised release and the use of the vulnerability assessment VI-SPDAT for triaging housing resources\footnote{The VI-SPDAT was rolled out a year after the Department of Housing and Urban Development imposed new requirements on cities receiving federal funding for housing to introduce centralized assessment processes (``coordinated entry systems"). VI-SPDAT was publicly available for free download. } \citep{codastoryWhosHomeless,orgcodeTimeSeems}. The extent of tradeoffs depends on the underlying correlations between baseline risk and efficacy, which should be explicitly investigated. For example, \citep{kube_community-_2023,kube_fair_2023,rahmattalabi2022learning,tang2023learning} study targeting housing allocations based on heterogeneous treatment effects under resource constraints, comparing equity-efficiency tradeoffs to the status quo. 

We outline these tradeoffs in the case of \textit{supervised release} programs in criminal justice, where RAIs are used to direct interventional resources (the supervised release program), though additional examples abound in other fields.

\begin{example}[Supervised release in criminal justice ]\label{ex:supervised-release}
    Supervised release is an intermediate release option besides pretrial detention (individual kept in jail before trial) and unconditional release. Supervised release is a term that refers to a broad array of programs that impose release with conditions in the probationary or pretrial settings. We focus on the pretrial setting, where supervised release can refer to everything from electronic monitoring programs \citep{kofman2019digital,doleac2018study} to supportive services programs with required attendance \citep{skemer2024comparing}.  
Its program design varies substantially across jurisdiction between additional sanction/surveillance or support. In Cook County, it comprises of electronic monitoring. 
 In New York City, Queens piloted a program with counselors and additional social services \citep{skemer2024comparing}. 
 
 The reasoning is that these conditions of release can enable releasing more individuals while reducing risk of re-offense (compared to unconditional release). Evidence is mixed on its efficacy.
\end{example}Understanding the institutional context is key to inform prioritization for supervised release. 
Pretrial detention can preclude re-offense by its nature, albeit at a large cost to civil liberties. On the other hand, supervised release is a much ``softer" decision and its effects on re-offense are unclear. How should judges determine who should receive supervised release?\footnote{There is some broad guidance that ``release conditions—if any are considered necessary—should be the least restrictive that reasonably assure court appearance and community safety," \citep{APPR-leastrestrictive} emphasizing $Y(D)$, the final outcomes of whatever release conditions are provided. Other guidance does indicate some preference for reducing risk of the riskiest individuals, suggesting that ``[limited community] resources should go to people with the highest need, meaning people assessed as the least likely to succeed."} \textbf{Heterogeneous treatment effects can introduce tradeoffs between these two competing objectives\footnote{Increased surveillance can increase frequency of technical violations of the supervision conditions and therefore worsen outcomes. Supportive services might work for some on the margin, but may not be enough to overcome other structural barriers. The limited evidence base is mixed, and variation in its program design makes generalization from prior studies even more difficult. 
}.}

To ground our discussion, we draw on data from Cook County to illustrate how different targeting strategies can lead to vastly different outcomes (see \citet{zhou2023optimal} for further details). For a period of time, the county used a decision-making matrix that recommended supervised release based on risk thresholds from the Public Safety Assessment (PSA), which predicts the likelihood of failure to appear (FTA) and new criminal activity (NCA) if released without intervention.

We can conceptualize this policy as recommending supervision to judges when the individual's predicted risk exceeds a threshold. As an exercise, we consider the outcomes when these recommendations are strictly followed. (In practice, judges have wide discretion). Formally, this corresponds to a rule of the form:
\[
\text{recommended supervision} = \mathbb{I}\left[ \E[Y^{FTA}(0) + Y^{NCA}(0) \mid X] > t \right],
\]
where $Y^{FTA}(0)$ and $Y^{NCA}(0)$ denote the potential risks of a failure to appear and new criminal activity under no intervention, respectively. This framing highlights how the existing ADS relies on predicted harms in the \textit{absence of treatment} to guide allocation of the \textit{intervention}.

\begin{figure}[t!]
    \centering
    \begin{subfigure}{0.35\textwidth}
\includegraphics[width=\textwidth]{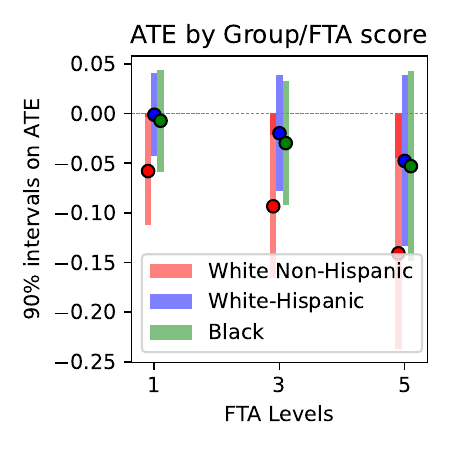}
        \caption{}
        \label{fig:superviserelease-ate}
\end{subfigure}\hfill\begin{subfigure}{0.55\textwidth}
        \centering
        \includegraphics[width=\textwidth]{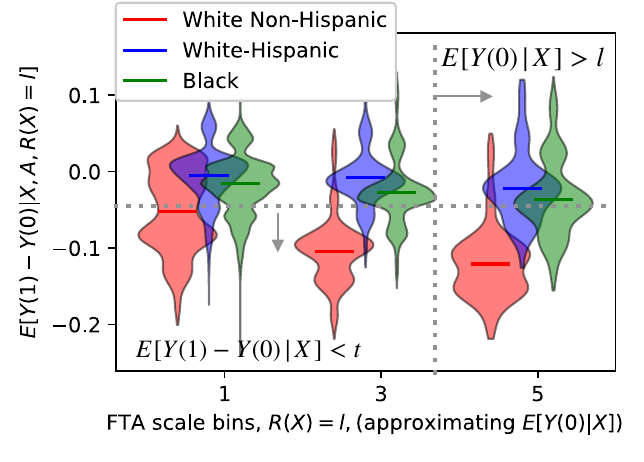}
        \caption{}
        \label{fig:supervisedrelease-hte}
    \end{subfigure}
    \caption{Estimated treatment effects of supervised release by race and FTA risk. (a) Average treatment effect (with 90\% confidence intervals) for different groups, stratified by FTA risk. (b) Distribution of heterogeneous treatment effects by group, stratified by FTA risk.}
\end{figure}

In \Cref{fig:superviserelease-ate} we show estimates of the ATE \footnote{Obtained by augmented balancing weights with a logistic regression outcome model and ridge balancing weights.} and 90\% confidence intervals, per race group (excluding an ``Other" category due to smaller sample size) and stratified by underlying FTA risk level. Although the average treatment effect is weakly decreasing (more effective at reducing FTA) for riskier individuals, there is substantial discrepancy in the absolute magnitude of this effect by race. There is also potential heterogeneity in treatment effects. \Cref{fig:supervisedrelease-hte} superimposes the distributions of estimated heterogeneous treatment effects (obtained by pseudo-outcome regression on the previous AIPW scores), again disaggregated by race and stratified on FTA risk level. Treatment effects are on average larger for white non-Hispanic than others (Black and white-Hispanic). There are many reasons this could be the case, such as different covariate distributions. 

The decision-making matrix thresholded its recommendations based on riskiness, as assessed via the PSA (\Cref{ex:justice}), in the absence of supervision. Although this can recall the riskiest individuals within races, and recall individuals with the largest treatment effects, this is at a targeting efficiency cost due to differences in absolute magnitude of treatment effects. The horizontal gray line in \Cref{fig:supervisedrelease-hte}
illustrates targeting based on heterogeneous treatment effects: choosing a threshold based on the difference that supervised release makes in reducing FTA. 

Although this re-uses already-implemented predictive risk systems, and is therefore readily implementable without additional research, it is only optimal for risk mitigation assuming that those at highest-risk for criminal re-offense are also those for whom supervised release is most effective at reducing re-offense.

These data explorations are not sufficient for making sharp policy prescriptions, as doing so would require deeper analysis of treatment effect heterogeneity and underlying mechanisms. However, given civil liberties concerns---particularly regarding disparities in supervised release and the disproportionate surveillance of communities of color---it is critical to examine equity-efficiency trade-offs more closely. What outcomes were prior decision rules optimized for, and what normative priorities did they reflect? If system designers believe that higher-risk individuals warrant greater scrutiny, this should be justified on moral grounds, not assumed under the guise of efficiency. \textbf{Ultimately, clarifying the moral and operational goals behind targeting choices is essential for designing interventions that are both fair and effective.}

Blurred lines between prediction and intervention are pervasive in other settings, beyond the examples we gave in the criminal justice system. 
 The following example of a \emph{benefit-based} intervention, drawn from \citet{black-smith-berger-noel-03-UI-rdd}, involves an ADS used to allocate capacity-constrained mandatory re-employment services and training for unemployment insurance recipients.

\begin{example}[Triaging social services, unemployment benefits] The Worker Profiling and Reemployment Services (WPRS) for unemployment insurance (UI) profiles claimants to estimate probability of exhausting their UI benefits (or expected duration on benefits), requiring mandatory employment and training services for claimants with high risk of exhausting benefits or longer duration on benefits. Ultimately claimants were scored from 1 to 20 (high risk). 
\end{example}
\begin{figure}
    \centering
    \includegraphics[width=0.5\linewidth]{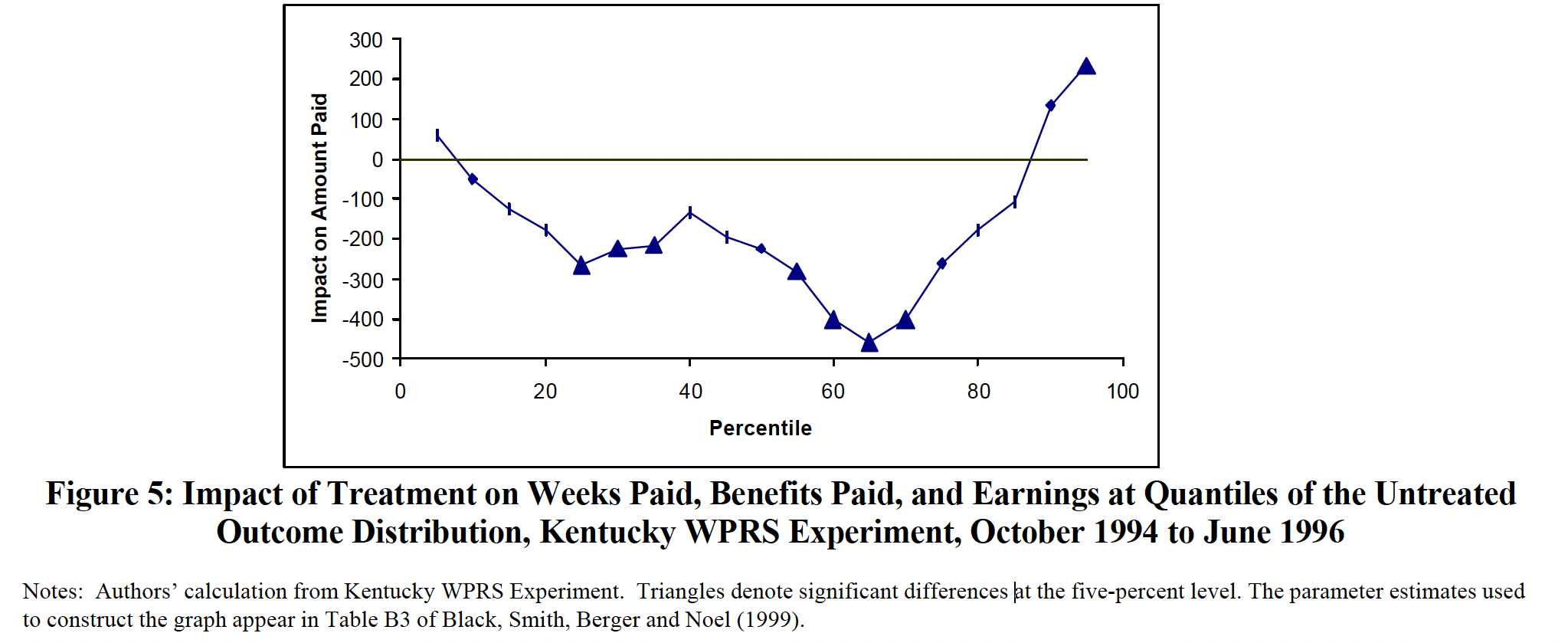}
    \caption{Figure from \citet{black-smith-berger-noel-03-UI-rdd}. Percentiles of baseline risk $\E[Y(0)\mid X]$ on the x-axis; estimated heterogeneous treatment effects on the y-axis. A nonlinear relationship indicates prioritization on baseline risk alone is suboptimal. }
    \label{fig:wprs-cross-calibration}
\end{figure}
We discuss details of the study in more detail in \Cref{sec:model-design-apx-wprs}. Even though the risk prediction model of \citet{black-smith-berger-noel-03-UI-rdd} improved predictions compared to prior models in other states, in part by using additional covariates and using a continuous rather than binarized outcomes, they find that \textbf{predicted duration is in general unrelated to heterogeneous treatment effects}. In this setting, different equity vs. efficiency objectives might favor different problem formulations. 

Other recent studies assess targeting based on baseline risk or need vs. welfare-efficient ITRs in different domains \citep{haushofer2022targeting,pmt-2016-brown,athey2025machine,sirota2025health}. 

Finally, we again acknowledge that there are also other important factors that justify targeting on $Y(0)$ rather than $Y(1)-Y(0)$, such as strong moral emphasis on targeting the worst off, alignment with intuitive or politically palatable conceptions\footnote{\cite{gowan2010hobos} studies homelessness and policy developments in San Francisco; the medicalization of homelessness received greater institutional support and influenced the development of vulnerability assessments that focus on mortality risk. In turn, these vulnerability assessments individualize homelessness and emphasize individual-level health factors rather than broader structural and systemic factors contributing to homelessness.}, or simply that it is simpler and statistically easier to estimate $\E[Y(0)\mid X]$ rather than heterogeneous treatment effects $\E[Y(1)-Y(0)\mid X].$

\subsection{Decision-theoretic Formulations}

In the previous section, we used real-world examples to illustrate how targeting decisions based on prediction alone can fall short of addressing institutional goals and tradeoffs. Here, we adopt a decision-theoretic perspective to conceptually distinguish when prediction is sufficient and when causal reasoning is necessary.  Intuitively, the basic distinction between these cases lies in how much causal understanding is encoded in the problem formulation itself and how much must be learned from the data. Before formalizing this idea, we give two parables:

\begin{parable}[\citet{kleinberg2015prediction}]
    First, imagine a person deciding whether to take an umbrella in the morning. It's enough to know whether it rains - the umbrella keeps them dry if it rains, and does nothing if it doesn't. Optimizing the decision problem reduces just to predicting whether or not it will rain. Meanwhile, taking an umbrella doesn't affect whether it rains. 
\end{parable}
\begin{parable}
    Consider a physician deciding whether or not to prescribe a patient a drug.  
    We would prefer to directly estimate the causal effect: whether the patient's outcomes will be better with the drug than without it. 
\end{parable}

Formally, the first case is where we are able to specify a loss function $\ell(D, Y)$ which depends on an observable quantity $Y$ which is \textit{not} itself a function of $D$, e.g., the decision to carry an umbrella does not itself impact whether it rains. The second case is where the appropriate loss function is parameterized by quantities that are not fully observable due to decision-dependence. For patients that we do not treat, we cannot observe their treatment effects, and vice versa.  

\begin{figure}[t!]
    \centering
    \begin{subfigure}{0.45\textwidth}
        \centering
        \begin{tikzpicture}[node distance=1.5cm]
            \node (D) at (0,0) {D};
            \node (Y) at (2,0) {Y};
            \node (L) at (1,-1) {$\ell(D, Y)$};

            \draw[->] (D) -- (L);
            \draw[->] (Y) -- (L);
        \end{tikzpicture}
        \caption{
        Can re-evaluate loss function on historical data under counterfactual decisions.}\label{fig:pureprediciton}
    \end{subfigure}\hfill\begin{subfigure}{0.45\textwidth}
        \centering
        \begin{tikzpicture}[node distance=1.5cm]
            \node (D) at (0,0) {D};
            \node (Y) at (2,0) {Y};
            \node (L) at (1,-1) {$\ell(D, Y)$};

            \draw[->] (D) -- (L);
            \draw[->] (D) -- (Y);
            \draw[->] (Y) -- (L);
        \end{tikzpicture}
        \caption{Causal estimation required. Corresponds to learning \cref{eq:itr} if $\ell(D, Y) = Y(D)$, i.e., loss/welfare is the potential outcome.}\label{fig:causal_loss}
    \end{subfigure}
\end{figure}

What are examples where prediction is enough? \citet{liu2023actionability} formalized a notion of a \emph{sufficient action} that improves outcomes for all decision subjects, regardless of their particular circumstances. Using a causal-graphical formalism, they show that the existence of such an action is both necessary and sufficient for pure prediction to be adequate for targeting interventions. 

\subsubsection*{Learning decision policies from observational data}

Different strategies for learning decision policies include ``plug-in prediction", plugging in prediction models $\E[Y\mid X]$ into expected loss, optimization over decision rules, or more advanced ``decision-aware learning" approaches that seek to learn predictive models that explicitly achieve the best decision-loss. An example of non-trivial decision-loss structure arises when cost minimization relies on global system state, such as shortest-paths, routing, and/or supply chain problems. See \Cref{sec:model-design-apx-decisionawarelearning} for more details.

\paragraph{Limitations and need for causal estimation} The approaches discussed so far leveraged our ability to evaluate the loss of \textit{any} decision for each historical example, allowing us to evaluate predictive models by the loss of the decisions they induce. The loss function encodes, via causal knowledge, how decisions relate to outcomes. 
These assumptions enabled us to select a fully observable target outcome $Y$ that suffices to specify the loss. 

In contrast, decisions impact outcomes in unknown ways (\cref{fig:causal_loss}), targeting decisions requires estimating HTEs, as we have discussed before. For example, if our goal is to optimize post-decision outcomes
    $\min_{\pi \in \Pi} \E[Y(\pi(X))]$,
then the implicit loss function $\ell(D, (Y(0), Y(1))) = Y(D)$ depends on both $Y(1)$ and $Y(0)$, which are never simultaneously observable.

\paragraph{Decision formulations: Specifying an objective function}
Although we have laid out basic distinctions between predictions and decisions, specifying the objective function and the constraints of the decision problem are also crucial. We refer the interested reader to \citep{barocas2023fairness} and references therein for discussions of additional desiderata, such as algorithmic fairness.\footnote{
The algorithmic fairness literature has developed tools for formalizing certain desiderata (such as group or individual fairness), and achieving them, often with constrained optimization. In practice, while auditing for unfairness has surfaced harms and issues with deployed predictive models, sources of unfairness and appropriate mitigations vary widely: perhaps an inappropriate target variable was chosen, the data ecosystem introduces statistical bias, the prediction only informs an institutional decision that is missing from the problem formulation, and so on. A central critique of considerations in the algorithmic fairness literature---that is closely related to the theme in this work---is the mismatch between constraints on model predictions and actual utility or improvements from decisions~\citep[see e.g.,][]{liu18delayed}. Other concerns include ``leveling down" objections to imposing fairness constraints that might worsen outcomes for one group while improving them for others (i.e. Pareto-suboptimal transfers of welfare) \citep{temkin2000equality,mittelstadt2023unfairness,hu2020fair,maheshwari2023fair}. 
} Other perspectives on algorithmic fairness emphasize consequentialist or welfarist accounts \citep{corbett2017algorithmic,corbett18measure,hu2020fair}, which are well aligned with our focus on evaluating realized impacts. Accurate and contextual measurement of realized impacts is a crucial input into any of these perspectives.

\subsection{Reconsidering Problem Formulation}
While we have outlined different technical formulations, how does one translate a real-world context into a technical problem formulation? We discuss important organizational considerations in problem formulation. Next, we discuss what kind of domain knowledge is required. Finally, even if prediction is sufficient, our prior discussions suggest opportunities to leverage interventional analysis to improve the design of predictive risk models.


\subsubsection{Ecosystem of Problem Formulation }


In data-rich settings, it can be tempting to instead take a ‘data-driven’ \textit{opportunistic} 
approach: exploring opportunities to innovate by training  candidate predictive risk models on available outcomes in the data, and proceeding further if a model achieves a high enough accuracy. Such an approach, however, is liable to produce algorithms that are ill-suited for the contexts in which they will be used \citep{coston2023validity}. Problem formulation should begin before looking at any data, involving key stakeholders in a conversation about the goals, success criteria, and potential risks of an algorithm in the proposed context \citep{kawakami2024situate}. Efforts to improve the model design and development process would benefit from an understanding of how data science teams currently navigate the model development process \citep{passi2019problem, passi2018trust,guerdan2025bricolage, kross2021orienting}.

The act of formalizing a problem and building a predictive model is in itself a policy intervention \citep{abebe2020roles}. Formalizing a problem can enforce clarity around goals, but introducing a quantitative frame for prediction can distort multiple goals by elevating and operationalizing some but not others \citep{green2021algorithmic,johnson2022bureaucratic,liu2023reimagining,elzayn2024}. 

A key lever is the choice of \textit{which decisions or interventions to target}. Causal analysis can identify key causal determinants of outcomes, and inform the choice of interventional frame. For example, \citet{lee2023causal} studied the causal determinants of length of hospital stay after cardiac surgery. Clinicians were interested in which factors --- such as medical staffing in the operating room --- could be changed to reduce length of stay. 




It can be extremely difficult to relate short-term, measurable outcomes to broader policy goals. Problem formulation also requires choosing target outcomes, which can have downstream ramifications on evaluation validity that we discuss in \Cref{sec:eval_science}.

\subsubsection{Domain and empirical knowledge is needed to justify data-driven targeting}\label{sec-model-design--justification}


Regarding the difficulty of predicting social outcomes, an important question is \textbf{how much can complex prediction improve upon simple prediction?} A mix of domain and empirical knowledge is required to answer this question, and social domains are often categorically different than other engineering ``signal recovery" domains.

\begin{example}[Social prediction, Fragile Families Challenge]
    In the ``Fragile Families Challenge", \citet{salganik2020measuring} repurposed a large-scale birth cohort study \citep{reichman2001fragile} into a common-task framework to assess the potential performance of machine learning models for social prediction (such as layoff, eviction, GPA) based on earlier survey responses. They found that while machine learning models could improve predictions, overall they displayed similar performance as simpler benchmarks designed with domain knowledge based on choosing just a few salient covariates. 
\end{example}

\citet{lundberg2024origins} probe further into the origins of unpredictability---interviews with participants reveal the crucial importance of unmeasured information in influencing their life outcomes. 

Multiple recent examples echo that simple predictions are competitive with complex machine learning for noisy social outcomes. Simple predictions should be standard benchmarks since they can be implemented for much lower total-cost-of-ownership. 
Recent studies find that much simpler predictors incorporating only neighborhood/district-level information \citep{perdomo2023difficult} or single-variable predictions \citep{stoddard2024predicting} are competitive with more complex prediction.\footnote{\citep{perdomo2023difficult} find that school and district-level information can target dropout interventions just as effectively as individual risk scores, reflecting the role of structural forces against the variability of individual variation. \cite{stoddard2024predicting} develop a predictive risk model to predict police misconduct, and that an exceedingly simple model -- number of past predictions -- is competitive (25\%-30\% relatively lower recall of top 5\% officers by estimated risk) with modern machine learning.} While ML culture may favor complex models, this can lead to ``worsenalization" \citep{suriyakumar2023personalization}, when additional covariates worsen predictive performance. 

The question of ``null efficacy" is also important for causal interventions. Establishing that treatments effects are heterogeneous before considering targeting is necessary. 
But, a null \textit{average} treatment effect could have opposite implications as to whether the intervention is overall ineffective, or whether there is underlying heterogeneity that was masked by averaging over subpopulations. It's folklore that in social domains, treatment effects are often noisy zeroes. 
It's harder to test for effect heterogeneity itself is harder than for a null average effect \citep{crump2008nonparametric,levy2021fundamental,sanchez2023robust}. 

\paragraph{Contextualizing the Relative Value of Prediction in Resource Allocation Problems}

Given that prediction is just one component of a broader system, there are many other ways of driving improvements in downstream social welfare apart from optimizing prediction quality. For example, directly expanding resource capacity and assigning interventions to more people, or just investing in increasing the effect sizes of the interventions themselves.

\citet{perdomo2023relative} and \cite{fischer2025value} study how do the welfare benefits of investing in prediction compare to those of other policy levers such as adding more resources. Through theoretical analysis and an in-depth case studied on unemployment prediction in Germany, they find that in highly resource-constrained regimes the impacts of adding more resources typically outweighs the marginal gains of targeting the same scarce resources better.

\subsubsection{Causal and Interventional Analysis Can Guide Predictive Task Design}

Even where pure prediction is enough, the ADS context in the social world requires greater care in the design of prediction tasks. The  interventional framework can be applied to improve prediction models, in a line of research sometimes called ``causal inference for machine learning". First of all, the embeddedness of our analyses in the social world, and consequent social dynamics, can affect the validity of model formulations. We discuss in more detail in \Cref{sec-apx-implicationsofsocialworldmodeldesign}.\footnote{To summarize, social interactions complicate predictive validity, for example via strategic, equilibrium or adversarial responses. Though these can be studied theoretically, it's important to consider fidelity of theoretical assumptions to real-world behavior, in order to avoid inadvertently reinforcing social inequalities by positing rational behavior as a standard. 
Social dynamics can also affect the validity of the causal inference framework most obviously via interference (outcomes affected by other individuals' treatment assignments), but in more subtle ways via the inherent conservativity of requiring well-defined \citep{stevenson2023cause} or modular \citep{hu2023race} interventions.} Next, even if pure prediction is sufficient,  interventional analysis can apply in different ways, which we discuss below. 


\textbf{Causal structure of data bias can inform variable selection }
Interventional analysis can express complex patterns that inform deliberations of including covariate information in a predictive model (or decision rule) or not.
One example is the question of whether protected attribute variables (such as race) should be included in a predictive model.
Some focus on the value that the addition of these variables may have for predictive accuracy \citep{manski2023probabilistic} depending on the use case \citep{basu2023using,coots2023racial} or for improving disparities in realized decisions. Insofar as race is understood to be a social category marking social difference (rather than, say, biological difference), the reasons that underlie race's predictive power in various social systems largely trace to those systematic social differences that define a racialized social structure. A world in which important life outcomes are racially stratified is a world in which race has powerful predictive power. The reasons for racial stratification today are of course the legacy of decades and centuries of race-making processes which differentially distribute(d) power, resources, rights, esteem, among other benefits and burdens along lines of perceived geographic ancestry or phenotype that we now call "racial". Causal graphical analysis, alongside other historical and social-scientific analyses, can be used to help draw out these complex correlation structures.

\textbf{Causal structure can express desiderata, such as anti-discrimination}

Interventional analysis, in terms of causal graphical models on covariates, can sharpen debates or highlight finer-grained desiderata. One recent important use was in algorithmic fairness. 
 Researchers have employed causal modeling to articulate how structural or systemic sources of unfairness and injustice surface in the data. More broadly, some causal dependencies among variables used in statistical analyses, including race and gender, might reflect historical and ongoing unfairness at the institutional or systemic level (in addition to inter-personal). By constructing simplified causal models of complex social systems, one can trace schematically “path-specific” causal effects representing various causal processes that generate the data we observe today. Path-specific effects can be fruitfully used to identify particular processes that generate structural injustices along race or gender lines \citep{kilbertus2017avoiding,kusner2017counterfactual,nabi2018fair,nabi2022optimal}. Concrete claims as to how certain race or sex causal effects are generated can help facilitate broader moral and political arguments for why certain causal dependencies should or should not be leveraged in prediction. 

That said, conceptualizing race and gender as causal variables has been controversial: for example, it's not obvious how to interpret counterfactuals (or causal effects) involving racial categories. 
Scholars have argued that these conceptual difficulties in defining causal effects of race imply different normative positions on what race is and how the social world works\citep{hu2020sex,hu2023race}. 
 Other work in causality-informed algorithmic fairness has suggested that causal definitions of discrimination can help to adjudicate what differences constitute discrimination on more formal grounds and hopefully clearer \citep{avoidingdisc17}. Other work has emphasized that among the potential benefits of using the causal formalism is the incentive to make ethical or normative (as well as empirical) commitments explicit and seek input from affected stakeholders, ethicists, and policymakers \citep{nabi2019learning}.


More generally,
the causal tools of path-specific effects can be helpful in learning predictors under constraints on these paths; see \citet{nabi2024fair,nabi2024statistical} for recent advances leveraging semiparametric statistics to include constraints on the infinite-dimensional predictor function.  

\textbf{Causal dependence structure clarifies interpretability, recourse, and other desiderata}

Concerns around learning shortcuts also motivate desiderata such as \textit{explainability} of predictive models. Causal models can be used to conceptualize different notions of dependence of \textit{prediction functions} on changes in inputs that have arisen in the interpretability literature, i.e. interventions on $X$ with impacts on a prediction function $\hat f(X)$ \citep{zhao2021causal,loftus2023causal}.

\section{Evaluation Science}\label{sec:eval_science}

Following the model design of predictive ADS, an equally important consideration is how to \emph{evaluate} whether these systems are working (or will work) as intended. In the context of ADS, \textbf{\emph{evaluation science} is the scientific process of quantifying the gap between articulated expectations for a system and the measured reality of that system's behavior or output} ~\citep{raji2022anatomy,mathison2004encyclopedia}. 

Multiple evaluation approaches provide meaningful evidence to assess claims about ADS.
A widely used approach for performance measurement is \emph{benchmarking} (Section \ref{sec:eval_perf_meas}), which evaluates the predictive statistics of the model underlying the ADS, such as accuracy, ROC scores, or F1 metrics on curated datasets, against a chosen proxy outcome. However, while useful, benchmarking provides only partial evidence for the performance of ADS, as high predictive accuracy alone rarely meets the real-world objectives of ADS deployments \citep{liu2023actionability,perdomo2023difficult}. There may also be concerns with the \emph{validity} of the evaluation (\Cref{sec:eval:valid}) due to \emph{measurement challenges} such as label bias and proxy-construct mismatch. 

This motivates an \emph{interventional} evaluation paradigm that moves beyond benchmarking to consider how predictions affect decisions and outcomes in real world deployments. In Section~\ref{sec:eval_exp}, we highlight different options for evaluating along the causal chain of $X\to R \to D \to Y$ as well as novel causal challenges that arise in doing so. 
Finally, in Section~\ref{sec:eval_broad}, we survey the broader social and organizational challenges in evaluating ADS, including the challenge of clearly articulating the goals and standards for an ADS system itself.
\begin{boxD}
\paragraph{Key takeaways} We argue that effective evaluation of ADS requires more than just measuring predictive accuracy. It involves assessing how predictions impact decisions and outcomes, and considering the broader social and organizational context in which the system operates. The evaluation process should include rigorous attention to measurement challenges, such as label bias and proxy-construct mismatch. Beyond benchmarking, an interventional evaluation framework based on randomized experimentation, causal estimation and modeling human-algorithm interaction offers a more comprehensive understanding of the true impact of systems in deployment.
\end{boxD}

\subsection{Benchmarking Failures for ADS}\label{sec:eval_perf_meas}
A critical part of the evaluation process is operationalizing policy goals into one or more \textit{target outcomes} ($Y$) to use for evaluation. To set the stage for a new framework of \textit{interventional evaluation}, we overview the standard paradigm of benchmarking. Standard benchmarking seeks to characterize the prediction error of the ADS (e.g. a risk score denoted $R$), against a target $Y$, using a static dataset. Predictive models might be compared on an alphabet soup of different error metrics. 
\citet{donoho2024data} attributes the success of data science to the ``frictionless reproducibility" of this exact benchmark-driven culture: shareable and open datasets, development of open-source and reproducible methodology, and competitive testing on shared benchmarks.

In many ways, ADS are \textit{not} developed in this paradigm. Vendors rarely disclose inputs or their algorithmic methodology under trade-secrets protection. Countervailing privacy considerations often prevent disclosure of consequential individual data. Improvements on benchmark test sets derived from ADS might rarely translate to improvement in the domain \citep{bao2021s}. 

\begin{casestudy}[Benchmarking Evaluation of an RAI]\label{casestudy-COMPAS-RAI}
Here we analyze a \emph{benchmark evaluation} of the COMPAS risk assessment tool~\cite{angwin2016machine}. 

Propublica replicated similar results as Northpointe for overall model accuracy, but found that although the \emph{error rates} between demographic populations were similar, Black defendants where the model was wrong were disproportionately given \emph{higher} scores. This indicates that the errors were distributed in a way that favored White defendants, by nudging judges to set a lower bail. In contrast, Black defendants with incorrect scores were likely to receive higher scores, which would nudge judges to be harsher. 
\end{casestudy}

So far, the evidence for and against the use of such risk assessments illustrates the broader dialogue on how to assess the impact of algorithmic interventions in high-stakes settings. We continue with case studies that highlight different approaches to evaluation, \textit{even for the same ADS and policy context}.

\subsection{Validity}\label{sec:eval:valid}
Validity concerns are pervasive in both benchmarking  and interventional evaluations.
The concept of \emph{validity} describes the extent to which an evaluation performed via one or more proxy outcomes captures the intended properties of a system \citep{coston2023validity}. A common reason why validity can fail in practice is that the evaluation is conducted under conditions which do not represent real-world deployment. This threatens external validity. We go into more detail about two pervasive issues: label bias, when available outcomes are proxies of ideal outcomes for evaluation, and selection bias, resulting in available data that systematically differs from the full population.

 

\subsubsection{Label Bias}\label{sec:eval:labelbias}

Label bias can arise from practical compromises when choosing outcomes in problem formulation. For example, a school might aim to assess whether an ADS improves the ``academic success'' of post-secondary students. A hospital system might evaluate the ``quality of care'' offered to patients. Yet these concepts are \textit{latent constructs} that are not directly observable in data \citep{jacobs2021measurement}. Instead, organizations use readily available \textit{proxy outcomes}, denoted $\tilde{Y}$ in \Cref{fig:Diagram}. These are stand-in variables such as post-secondary graduation (for academic success) or hospital readmission (for quality of care). A \textbf{proxy-construct mismatch} arises when such proxy outcomes only imperfectly capture the underlying target construct \citep{obermeyer2019dissecting, mullainathan2021inequity, neil2024algorithmic}.

\paragraph{Consideration of construct mismatch in proxy outcomes is essential in evaluating ADS' impact.} 
Proxy outcomes used to evaluate an ADS may imperfectly reflect broader policy goals. This measurement gap raises key evaluation challenges: How do we select the most relevant proxy outcomes to track? 
How do we characterize potential gaps between proxy outcomes and ideal targets?
Should we attempt to collect additional data? 

As illustrated by the following two examples, overlooking proxy-construct mismatch can yield misleading evaluations of RAIs:

\begin{issue}[Proxy Outcomes in Healthcare, c.f. \Cref{ex:medical_cost}]
A widely deployed healthcare enrollment algorithm was evaluated for racial bias using healthcare costs ($\tilde{Y}$) as a proxy for patients' health needs ($Y$). An initial evaluation performed via healthcare cost suggested the algorithm was well-calibrated across racial groups. However, healthcare costs proved to be a poor proxy for the target construct of healthcare needs: Black patients with similar health needs to white patients typically incurred lower costs due to structural barriers in access to healthcare. When researchers re-evaluated the algorithm using more direct measures of healthcare needs, they uncovered significant racial disparities in the algorithm's predictions \citep{obermeyer2019dissecting}. This example illustrates a specific form of proxy-construct mismatch called ``label bias.''
\end{issue}

\begin{issue}[Label bias in Criminal Justice, c.f. \Cref{ex:justice}]
In criminal justice, predictions of new criminal activity are often evaluated using proxy outcomes ($\tilde{Y}$) such as re-arrest or failure to appear in court. However, re-arrest and failure to appear may poorly reflect the target construct of actual criminal activity due to reporting practices and overpolicing. When evaluating an RAI's impact, relying on re-arrest as a proxy for re-offense can yield misleading assessments of predictive efficacy of new criminal activities~\citep{neil2024algorithmic}.
\end{issue}
In the case of arrest and offending, this problem has long been known in criminology \citep{biderman1967exploring}, and scholars have suggested various ways to think about dealing with this data mismatch \citep{zilka2023progression, neil2024algorithmic,nagaraj2025regretful,schoeffler2025indeterminacy}. The validity of proxy outcomes can break down in two key ways: sometimes the target construct is fundamentally unobservable (like true criminal behavior \citep{zilka2023progression, neil2024algorithmic, biderman1967exploring}), while in other cases the target construct could be measured but practical barriers prevent its collection (like detailed health needs assessments). 

Several strategies can bolster validity of conducting evaluations with proxy outcomes, such as assessing against \textit{multiple} proxy outcomes (e.g., re-arrest, failure to appear) \citep{kleinberg2018human, Arteaga18selectivelabels}. Demonstrating that an RAI evaluation holds under different plausible assumptions on the relationship between target and proxy outcomes (i.e., sensitivity analysis \citep{fogliato2020fairness}) provides evidence for validity. 


 \subsubsection{Selection bias from prior decisions }

Throughout, we've emphasized the important role of decisions made by organizations in achieving better outcomes. Introducing these decisions into a default conceptualization of ADS implies that they were also present in historical data used to train predictive models. These decisions can be understood in the framework of observational causal inference: when they induce censoring, this has often been described as ``selective labels" \citep{lakkaraju2017selective}, where ``observed outcomes are themselves a consequence of the existing choices of the human decision-makers".

This issue arises in bail decisions, a slight variant of the PSA setup in \Cref{ex:justice}. Setting bail is an additional decision in between detention without bail or release without conditions. Individuals who are offered bail may be unable to pay it, and then detained pretrial\footnote{ The defendant posts (pays) bail (if it is a secured bond), or not. If an individual cannot post bail, in the sense that they cannot arrange funds to pay the secured bond, they may be detained pretrial. If an individual makes bail, they are released into the community (albeit with a financial incentive to appear for trial), and all outcomes are possible: they may make their trial or commit another crime. }. If an individual is detained pretrial $(Z=1)$, they definitionally cannot \textit{fail} to appear for their trial and their ability to commit another crime is severely curtailed; their downstream outcomes are censored. Therefore, changes in bail policy introduce distribution shifts --- the New York Criminal Justice Agency's validation of the release assessment \citep{nycja-2025validation} finds that average risk of released individuals increased after bail reform resulted in releasing significantly more individuals.


From the perspective of the individual who was detained pretrial, if they were released \textit{instead}, they \textit{may or may not have} re-appeared for trial (1-FTA), or committed another crime (NCA). This is a \textit{counterfactual} that depends on joint potential outcomes --- what would have happened if an individual was released instead. Contrast this with a societal consequentialist perspective that assigns utility or cost to final outcomes (failure to appear or committing another crime) and decisions (societal cost of bail/detention) marginally --- viewing detention as \textit{equivalent} to reducing failure to appear. 
The individual counterfactual perspective resonates with moral debates whether or not cash bail ``punishes the poor" or otherwise improves outcomes via ``specific deterrence".\footnote{More specifically, these arguments posit that cash bail ``punishes the poor" by causing detention for poor individuals unable to afford bail bond who otherwise would not have offended. Or, they argue on the contrary that money bail provides ``specific deterrence" in that it provides financial incentives for an individual to avoid failure to appear relative to if they had in fact been released on recognizance, without any conditions.} But, it is not identified from observed data and requires additional assumptions or structure. Nonetheless, the contrast highlights that \textit{selective labels censor valuable information and introduce distribution shift challenges}.

The problem of selective labels is pervasive beyond criminal justice alone. 
For example, banks do not observe loan default or payback outcomes for individuals they deny loans for.\footnote{In lending, they study methods for addressing this under the terminology of ``reject inference" \citep{banasik2007reject,ehrhardt2021reject}.} Therefore benchmarking evaluations on available data are assessing, for example, error of a predictor $\E[(Y-\mu(X))^2 \mid Z=1]$ on the censored-in population\footnote{This selection bias poses a key challenge for external validity: researchers found that evaluations of credit-lending algorithms on a sample of approved applicants systematically misrepresents the accuracy and fairness properties on the full population of loan applicants due to distribution shifts \citep{coston2021characterizing,chien2023algorithmic}. }. 
In some settings, institutions can take advantage of randomized samples to avoid selective labels. The IRS, for instance, maintained a ``National Research Program,'' which is a stratified random sample of taxpayers who are audited, which can help overcome selective labels \citep{elzayn2024}. But the presence of such representative samples with labels of interest can be rare in resource constrained settings.

\paragraph{Conclusion and takeaways}
In practice, it's rare to measure ideal outcome and/or to have a representative sample from the target population for evaluation. Given the validity challenges we have outlined above, it becomes necessary to reason about how well available outcomes and available data samples approximate the ideal ones. If relevant, statistical methods can correct for common data problems such as measurement error \citep{guerdan2023counterfactual} and selectively missing outcomes or data \citep{coston2020counterfactual,Kallus2018Residual}. These often leverage methodology for missing data, which is quite closely related to causal inference and interventional analysis. 

 \subsection{Towards an Interventional Evaluation Paradigm}\label{sec:eval_exp}

We now turn from benchmarking to \emph{interventional evaluation}: studying how predictive systems causally affect outcomes.  
Formally, an ADS produces a risk score $R$ that influences decisions $D$, which in turn affect outcomes $Y$.  
Different empirical settings reveal different parts of this chain.  The appropriate evaluation design depends on both
(i) \emph{experimentation capacity}—whether randomized assignment of decisions or ADS access is possible—and
(ii) \emph{information availability}—whether we observe individual-level $(R,D,Y)$ or only aggregate outcomes.

\paragraph{Hierarchy of evidence.}
Randomized controlled trials remain the most credible way to establish causal effects, yet they are often infeasible in high-stakes institutional settings.
Rigorous evaluation therefore meets organizations ``where they are,'' combining designs based on retrospective observational data, natural or quasi-experiments, and prospective randomized rollouts.  In practice, robust evidence may require both pre-deployment (off-policy or sandbox) analyses and post-deployment field trials.

\begin{figure}
    \centering
    \includegraphics[width=0.5\linewidth]{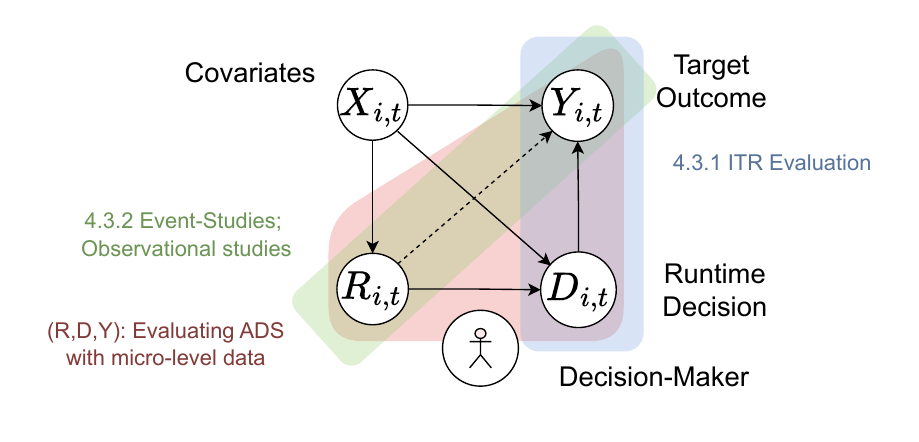}
    \caption{Our interventional evaluation paradigm vs. the status quo of ``blind men studying the elephant", highlighted.  Blue: $D\to Y$ (ITR evaluation).  
    Green: policy-level impacts of introducing an ADS (event studies).  
    Red: $R\to D \to Y$ with micro-level data (human--algorithm interaction and encouragement designs).}
    \label{fig:eval-science}
\end{figure}

\paragraph{Our integrated framework: $X\to R \to D \to Y$.} We emphasize the causal chains between predictions, interventions and outcomes. To organize the growing literature, we decompose evaluation efforts by the links they estimate. We begin by overviewing methods that assess the causal impact of decisions or interventions on outcomes {$D \!\to\! Y$,} such as evaluating an individualized treatment rule or targeted policy.
Other approaches for the \textit{effects of predictions} appeal to the presence of some policy change that introduces ADS: based on macro-data alone, some approaches estimate \textit{aggregate effects} of introducing or withdrawing ADS (e.g., via event studies or difference-in-differences). Elsewhere, other studies focus on \textbf{$R \!\to\! D$:} studying how human decision-makers rely on or respond to risk scores. One case that simplifies estimating the effect of $R\to D \to Y$ is if the decision $D$ is a function of the risk score; else, there are fundamental causal challenges. We overview different approaches. Altogether, viewing these disparate perspectives in our integrated framework provides a unified toolbox for \textit{understanding the impacts of ADS on social welfare}.

\subsubsection{Evaluating ITR performance is a problem of off-policy evaluation}\label{sec-evalscience-itr}

\begin{wrapfigure}{l}{0.25\textwidth}
\includegraphics[width=0.3\textwidth]{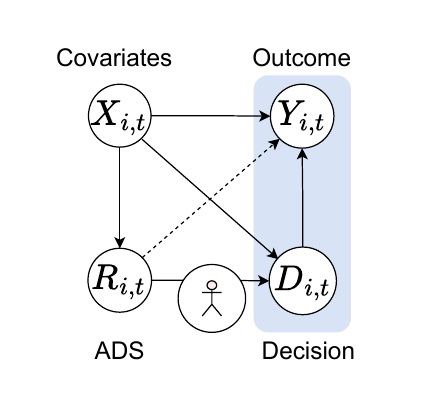}
\end{wrapfigure}

To highlight the challenge of identifying and estimating performance measures in an interventional framework, we begin with an overview of how defining causal equivalents of performance measures those discussed in Case Study \ref{casestudy-COMPAS-RAI} (COMPAS and classification errors), introduces different causal estimands. For example, consider generalizing a fairness measure like true positive rate to assessing impacts of interventions on binary outcomes. The fundamental problem of causal inference introduces statistical challenges: if we care about outcomes under alternative decisions, they were not observed, and therefore cannot be directly audited from data. Therefore, almost any performance measure that might be a ``standard" metric in the pure prediction case, needs to be translated for the interventional setting.

Disparities in true positive rates $P(\hat Y=1 \mid Y=1, A=a)$ measure differences in receiving positive classifier decisions $\hat Y$ conditional on having the underlying ``positive label" such as not defaulting on a loan (in lending prediction) or re-appearing for trial (in RAI prediction). 
One way to define an analogous \textit{interventional} true positive rate is to posit that an individual should have received a \textit{positive decision $D=1$} if in fact they \textit{would have improved if they received the intervention}. 
The interventional analogue posits that those who strictly benefit, i.e. \textit{respond} under the intervention $Y(1)>Y(0)$ \citep{kallus2019assessing}, ought to receive the intervention, $D=1$. \citet{imai2023principal} also develops fairness criteria based on these principal strata. A true positive rate disparity across groups can be defined as follows.  
\begin{equation}
    P(D \mid Y(1) > Y(0), A=1) -  P(D \mid Y(1) > Y(0), A=0)  \tag{Responder TPR disparity}
\end{equation}
Importantly, the fundamental identifiability challenge is that both potential outcomes $(Y(0),Y(1))$ are never jointly observed, and neither is the event $\{Y(1) > Y(0)\};$ it represents an unobserved \textit{principal stratum}. An additional identifying assumption of \textit{monotonicity}, that $Y(1) \geq Y(0)$, or that the intervention only improves outcomes, can restore identifiability. \citet{kallus2019assessing} give bounds under potential violations of monotonicity.

 \subsubsection{Evaluating ADS with quasi-experimental and observational 
 causal analysis 
}\label{sec-eval-RDY-obs}

So far, we have highlighted the causal difficulties of evaluating the impacts of different interventional decisions $D$ alone on outcomes $Y$, without addressing  how providing ADS ($R$) affects these downstream decisions and outcomes $D,Y$. Different models of the relationship between $R,D,Y$, including actual or statistically plausible randomization of the provision of ADS via $R$ or in decisions $D$ imply different evaluation techniques. 


\paragraph{The introduction of an ADS is a policy intervention: macro-data and event-studies}

\begin{wrapfigure}[12]{l}{0.25\textwidth}
\includegraphics[width=0.3\textwidth]{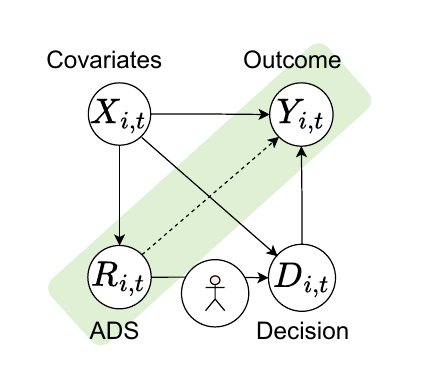}
\end{wrapfigure}
One way to model the causal impacts of ADS is to view the introduction of the ADS itself as a policy intervention or ``program", considering the impacts on relevant downstream outcomes prior vs post the introduction of an ADS tool into practice. Often these studies use event-study pre-post designs on outcomes before and after the introduction of the ADS. 

An example of such a study is \citet{albright2019if}, which investigated the impacts of the introduction of ADS for pretrial risk assessment. 

\begin{casestudy}[Before-after comparison with panel data and quasi-experimental methods]\label{sec:casestudytwo}

\cite{albright2019if} assesses observational case-level data sourced from Kentucky before and after House Bill 463 (HB463)
which required judges to more actively factor in the the Kentucky Pretrial Risk Assessment (KPRA)'s risk label (high, moderate and low) in their decision-making. They were requested to release all low and moderate risk defendants with a non-financial bond unless otherwise justified.

The author first analyzes temporal trends and identifies
a discontinuity in the percentage of non-financial bonds after the policy implementation. They conclude that considering the ADS 
overall leads to more leniency in judge decision-making, the scores have little to no effect on decisions on high-risk defendant decisions of any race, including black. 
A difference-in-difference analysis reveals that pre-HB463, differences between Black and White defendants are negligible. However, post-HB463, white defendants receive a significantly greater amount of leniency, experiencing 30\% and 62\% more of the short-term gains in non-financial bonds than their Black counter-parts in low and moderate risk score settings. Even when adjusting for time-dependent judge-fixed effects
Black defendants still experience a 25\% lower increase in non-financial bond issuance. 
\end{casestudy}

Event-studies are often used to study the impacts of non-randomized broad policy interventions. Broad policies can have complex relationships with specific ADS and decision-making patterns. 

Such studies based on before-after contrasts of outcomes can measure aggregate impacts of introducing ADS at all\footnote{\citep{zhou2021empirical} conduct an event-study analysis of initial bail reform efforts in New York; but the impacts of bail reform on the different distribution of released defendants are complex.}. However, this type of event-study analysis typically doesn't consider the per-individual information of how the ADS predictions specifically impacted decisions. 
Evaluating ADS requires understanding both the magnitude of their overall effectiveness, and disaggregated along the potential mechanisms behind their impact. 
Implications for next steps could be vastly different: should an organization investigate what informs front-line decision-making in more detail, or should they develop more effective programs/interventions to allocate? 
We next discuss different evaluation frameworks for disaggregating evaluations based on individual-level micro-data decisions and outcomes. 

 \subsubsection{Evaluating human-decision-makers' use of ADS}
 
\paragraph{Challenges in evaluating AI-assisted human decisions}\label{eval-sub-section-ai-assist-human}
The complexity of intervening in sociotechnical decision systems requires a more nuanced approach than traditional RCTs. ADS evaluations must account for the decision-making process ($D$) influenced by algorithmic predictions. Understanding interactions between risk scores, human decision-makers and their final decisions is crucial. 

\begin{wrapfigure}[10]{l}{0.25\textwidth}
\vspace{-9mm}
\includegraphics[width=0.3\textwidth]{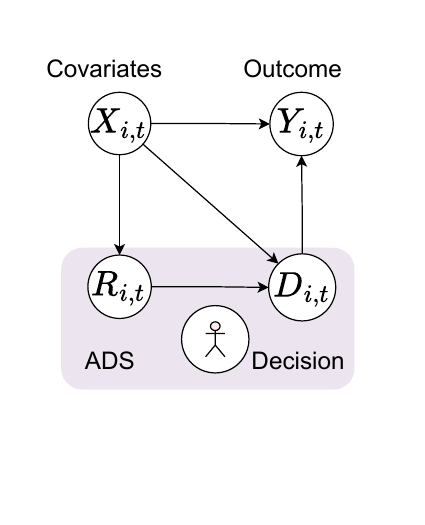}
\end{wrapfigure}

When it is not possible to randomize assignment of a prediction system to human decision-makers, a common evaluation strategy when decisions have known effects is to analyze observed human decisions that were informed by a predictive model to infer whether the human is appropriately relying on the model. But how does one define \textit{``appropriate" model reliance}? 
A recent survey on approaches to measuring reliance \citep{passi2022overreliance} describes methods based on counting acceptance of AI recommendations (out of all recommendations) or switches to match AI recommendations (switch fraction) or derived metrics (weight of advice) \citep{buccinca2021trust,lu2021human,kim2023algorithms,logg2019algorithm}. But metrics that measure ``incorrect switches" based on ex-post realizations of probabilistic outcomes
can be misleading, because they do not account for the limited information available to the decision-maker at decision time. A decision-maker may go with their own judgment because the ex-ante probability that they are correct is higher than the probability that the model is correct, but still appear to make the ex-post wrong decision because the outcome is not deterministic.  Hence, any definition of appropriate reliance must account for the limited information available to the decision-maker at the time of the decision.

Such a definition can be derived within a decision-theoretic framework similar to that discussed above, as the rate at which an idealized decision-maker with access to the human's independent predictions and those of a predictive model relies on the model. Consider a rational Bayesian agent who sees the independent human judgment and the model prediction as part of the ``signal'' available at the time of the decision (in addition to features $X$, model explanations, and other provided information), and must choose which to go with as the final decision. The ideal agent knows the prior probability of model correctness and the joint distribution over signals and states: they can then conduct Bayesian posterior updating and take belief-optimal actions \citep{guo2024decision}.
The performance of this agent benchmarks the maximum attainable performance of a human deciding between their own and the model’s prediction. \citep{mclaughlin2022algorithmic} study a rational decision-making perspective with preferences, while \citep{mclaughlin2024designing} leverages the potential outcomes framework. 

The expected performance of the maximally-informed, Bayesian rational agent provides a means of estimating the potential for a human and predictive model pair to attain ``complementary performance,'' where the performance of the AI-assisted human exceeds that of either agent in isolation. The maximum attainable performance of the rational agent with access to human and model predictions can be compared to the best attainable performance of the rational agent constrained to making the best (fixed) choice between the two given only the prior \cite{guo2024decision}, capturing the expected ``boost'' in decision quality from having access to both agents' decisions. 
To do so requires independent human and AI decisions for a set of representative instances.

Beyond assessing the potential for complementary performance and benchmarking how close a human-AI pair comes to the upper bound on performance, we are often interested in how and how well agents (human and artificial) exploit available information. 
A longstanding, robust finding in decision studies is that when statistical prediction is evaluated head-to-head against clinical (i.e., human prediction), the statistical model typically exceeds or matches the human experts' performance \cite{aegisdottir2006meta,grove2000clinical,meehl1954clinical}. This finding goes against common intuitions that human decision-makers can possess valuable private information that is hard to encode in statistical models, such as through interacting directly with the decision subject. Recent works explore the presence and value of information uniquely available to humans or other agents \citep{alur2024auditing,Schechtman2025discretionloophumanexpertise,guo2025value}.

It's crucial to understand the relationship between $R$ and $D$ since often human decision-makers oversee final decisions in domains whose consequence precludes randomizing final decisions. This relationship affects how changes in predictive model characteristics translate to changes in final decisions (and therefore outcomes).

\paragraph{The impacts of predictive models more generally: micro-data in the observational setting.}

So far, we have discussed how to study causal effects of decisions on outcomes ($D$ on $Y$), introducing ADS on outcomes with macro-level data ($R$ on $Y$) as well as evaluations of human decisions made with ADS ($R$ on $D$). Next, we discuss how to use micro-level data on $(R,D,Y)$ to assess impacts of $R$ on $Y$ with varying identifying assumptions. 

\begin{wrapfigure}[10]{l}{0.25\textwidth}
\vspace{-9mm}
\includegraphics[width=0.3\textwidth]{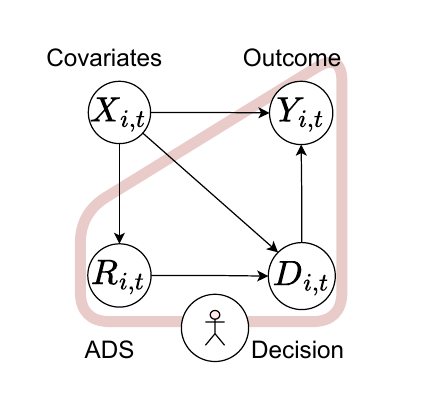}
\end{wrapfigure}

\paragraph{When prediction models determine decisions}

For simplicity, we first consider a case when algorithms set the boundaries of who does or does not receive a decision, deterministically. 

\begin{equation}
    D = g(\hat R), \qquad \hat R \in [0,1]
\end{equation}

This tends to be the implicit assumption in binary classification, i.e. that predicting default implies giving loans to those predicted not to default (and that those individuals would accept). For example, binary classification might consider thresholds $D = \mathbb{I}[\hat R > \frac 12]$, when $\hat R$ is a probabilistic risk score, $\hat R \in [0,1].$

A popular tool is to use ideas from \textit{regression discontinuity design} to leverage the deployment of a predictive algorithm for the purposes of evaluating the causal impacts of that decision \citep{perdomo2023difficult,gallo2024effectiveness,cowgill2018impact}.\footnote{A closely related design is the \textit{tie-breaking design}: capacity constraints that might result in randomization at the final threshold of eligibility. For example, \citet{smith2002threat} is an early use of this design to evaluate the impacts of ``profiling" unemployment insurance seekers by their risk of lengthy unemployment spells and targeting them for mandatory re-employment services. 
} 

Since predictive risk models rarely themselves impact final outcomes, since they often only inform decisions, ``fuzzy RDDs" are typically more appropriate. 
 Although RDDs are generally compelling, causal effects are only defined locally for individuals with covariates right at the threshold. 

\paragraph{Observational analysis of $R$ as a treatment introduces overlap issues.}

When ADS informs decisions without dictating them, conceptual and technical challenges arise in formalizing: \textit{what is the causal effect of a prediction model?}. A key challenge is that probabilistic risk predictions are often deterministic functions of covariates, i.e. $\hat R = f(X),$
for some function $f$ such as logistic regression, a decision tree, etc. Therefore considering $\hat R$ as a treatment violates the overlap assumption: causal inference requires some randomness in treatment assignment, to be able to generate counterfactual contrasts of similar individuals who nonetheless receive different treatments. See \citet{mendler2022anticipating} for further discussion of the challenges of direct causal effects of $R=f(X)$ on $Y$. Some approaches use robustness around model predictions: \citet{ben2021safe,zhang2022safe}, in the face of violations of overlap, consider robust extrapolation based on structural and shape assumptions (i.e, Lipschitzness or other smoothness assumptions) for evaluating optimal treatment allocation without randomized (or quasi-randomized) assignment. 

On the other hand, we have discussed separate conceptualizations of effects of $D$ on $Y$, and $R$ on $D$, which gives intermediate tools to combine with the analysis of individual-level data. Another approach is to separately model impacts of predictive models on outcomes of interest, for example via separate decision/interventions. \citet{zhou2023optimal} studies impacts of algorithmic recommendations as encouragements into treatment, which formally imposes an covariate-conditional exclusion restriction by assuming that algorithmic recommendations affect outcomes only via changing final decision/interventional assignments, only the latter of which has direct effects on final outcomes. If there is enough treatment variation, conditional on covariates, this implies that robust extrapolation is only needed for the treatment response to recommendations. 

\paragraph{Human-decision makers introduce unobserved confounders}

\begin{wrapfigure}[10]{l}{0.35\textwidth}
\vspace{-9mm}
\includegraphics[width=0.35\textwidth]{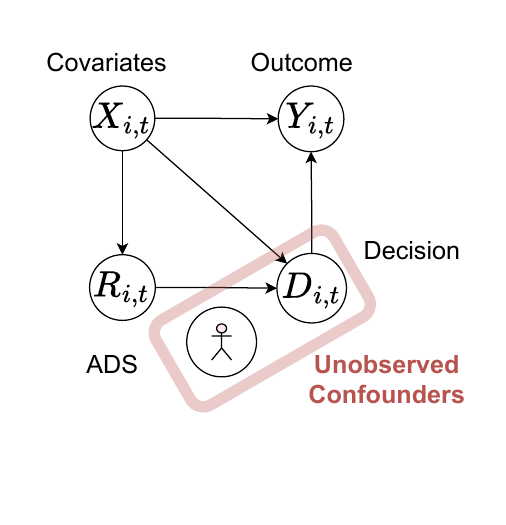}
\end{wrapfigure}

The presence of human decision-makers in institutional context implies there are probably \textit{unobserved confounders}, such as caseworker discretion or information about decision-subjects,  that are not recorded in limited administrative data. Unobserved confounders affect prior decisions and outcomes and pose fundamental challenges to causal inference, as observed differences in outcomes could be driven by differences in unobserved confounders. 
Unfortunately, the observational joint distribution is not informative of the presence of unobserved confounders: these need to be reasoned about with domain knowledge or auxiliary information/structure like instrumental variables or proxies. Sensitivity analysis and partial identification methods can instead bracket the range of what can be justified from observational data, subject to restrictions on how strong the unobserved confounders should be \citep{kallus2021minimax, guerdan2024predictive}.

\subsubsection{Experimental and in-deployment evaluations}

So far, we have discussed a whole range of strategies to use in observational data. 
In causal inference, the strongest evidence is typically considered to be a randomized controlled trial (RCT). 
Introducing random variation in whether a decision instance receives ADS information may be the ideal RCT.


Designing and implementing experimental evaluations for ADS in view of these complexities is an open area of research. The next case study outlines a RCT of provision of ADS (the PSA) and an example of interventional evaluation via principal stratification.
\begin{casestudy}[{ Experimental Evaluation}]\label{casestudy-experimental}

During a 30-month assignment period, spanning 2017 until 2019, judges in Dane County, Wisconsin were either given or not given a pretrial Public Safety Assessment (PSA) score for a given case. 
Although the PSA was calculated for every case, randomization determined whether the judge saw the score ~\citep{imai2020experimental}. Given these scores, the judge made the decision $D_{i,k}$ to either enforce a signature bond, a small ($<$\$1000) or large case bail ($>$\$1000). Information about judge decisions, and defendant outcomes, $Y_{i,k}$ (ie. failure to appear (FTA), new criminal activity (NCA) and new violent criminal activity (NVCA)) are tracked for a period of two years after randomization. 

The causal question is in determining how the treatment of providing PSA scores for a given case impacts the decision being made by the judge, and the supposed \emph{correctness} of that decision. The authors state that it should be optimal for the judge to be more lenient for lower risk cases, and harsher for higher risk cases and stratify the treatment effect analysis by risk level. They develop the following average principal causal effect (APCE) for various risk levels:

$$ APCE = \E[ D_i(1) - D_i(0) | Y_i(1) = y_0, Y_i(0) = y_i]$$

Where $ D_i(1)$ is the decision under treatment (showing the PSA) $Z_i$ and $Y_i(1)$ are the outcomes under treatment. $D_i(0)$ and $Y_i(0)$ mean $Z_i = 0$ and these are the results under no treatment. $y_0$ and $y_i$ are $\in \{0,1\}$. When $(y_0,y_i) = (0,1)$, then it is a preventable risk, meaning for example, an NCA occurs if released. When $(y_0,y_i) = (1,1)$, the outcome would happen regardless of release and when $(y_0,y_i) = (0,0)$, the outcome would not occur regardless of release. 

The main empirical result is that when looking at average treatment effects on judge decisions and downstream outcomes, there is a negligible impact of the PSA provision on the judges' overall decision-making. As a result, there were also not many notable differences in downstream outcome.  

Interestingly, the authors conclude that there are meaningful demographic differences with respect to gender. Making the PSA visible leads to more lenient decisions for female defendants, while male defendants are relatively unaffected. The PSA provision does not seem to have a significant disparate impact due to race. 
 \end{casestudy}

\subsection{Broader Social Challenges in Evaluating ADS}\label{sec:eval_broad}

Although we've focused so far on conceptual and methodological technical challenges with formulating evaluations of ADS, the social and organizational context of ADS introduces additional challenges for evaluation. 

\subsubsection{Defining ``Better" Decision-Making in Evaluations }
So far, we have emphasized singular choices of outcomes along which to assess whether ADS leads to improved decision-making. However, what constitutes a ``better" decision or decision-making process is often underspecified and contested. As we have observed throughout, ADS address complex social issues without clear solutions, affecting stakeholders with diverse interests and commitments. Many institutional decisions, especially in high-stakes public-sector contexts, involve multiple, sometimes conflicting, higher-level objectives \citep{green2021algorithmic, zacka_when_2017}.\footnote{For instance, college enrollment algorithms that rank applicants for admissions or financial assistance might be evaluated according to their original purpose -- often, optimizing yield -- but good college admissions decision-making may serve additional goals valued by stakeholders, such as ensuring student academic success or enhancing campus diversity \citep{engler_enrollment_2021}.} 

Beyond who gets selected, ADS may also be evaluated in terms of procedural and operational concerns, such as how resource-intensive a decision-making process is or how legitimate the system appears to key audiences \citep{grgic-hlaca_dimensions_2022, martin_are_2023, yurrita2023disentangling}. Different stakeholders may weigh these evaluative dimensions differently, making it difficult to establish a single, universally accepted benchmark for ADS \citep{nguyen2024definitions, johnson2025predictive}. A narrow evaluation framework that privileges only one of these dimensions -- such as predictive performance -- risks obscuring critical welfare trade-offs. 

\paragraph{Connection to Standards Development in Program Evaluation}

In the field of program evaluation, there is a major emphasis on the development of performance standards. The proper formulation of evaluations in the context of a machine learning task,  including the choice of outcomes to measure, presents a similar challenge.
 In particular, in many legal product safety regimes, the clear articulation of system expectations is essential in setting up robust audits and performance verification schemes~\citep{raji2022fallacy, raji2022outsider}. This is also closely connected to performance management. Under the U.S. Government Performance and Results Act, for instance, agencies must define and report performance measures. Many agencies institute quality assurance programs -- often utilizing random samples of cases -- to measure and improve outcomes \citep{ames2020due}.

\textit{Internal standards} are the starting point via internal auditing and pre-deployment testing~\cite{raji2020closing}, articulating organization expectations to inform local procurement and model testing. Proprietary standards include those developed by standards-making bodies, such as International Organization for Standardization (ISO), or the Institute of Electrical and Electronics Engineers (IEEE), or other widely disseminated guiding documents such as NIST’s Risk Management Framework (RMF), Microsoft's AI Fairness Checklist or Google's Model Cards~\cite{mitchell2019model}, widely disseminated across a range of organizations~\cite{ojewale2024towards}. In many cases, the expectations of the system are set by law in \textit{governmental standards}. This can include, for example, fairness criteria set by legislative anti-discrimination statues ~\citep{xiang2020reconciling,wachter2020bias,xiang2019legal}, or product safety standards of performance set by rule-making enforcement agencies ~\cite{raji2022fallacy}.

Beyond general standards for the system, the institutional context or change generates other challenges around evaluating the performance of an AI-assisted human decision-maker. These include how to “onboard” the human so that they form a reasonable expectation about how the model will perform, how to ensure that model updates do not decrease performance by violating the human’s expectations \cite{bansal2019updates}, and how to understand the extent to which the human’s beliefs about the true data-generating process are shaped through use of the AI or model more complex sequential decisions where predictions inform future information gathering or sub-actions.

\subsubsection{Institutional Inter-dependencies, Spillover Effects, and Bounding the Scope of ADS Evaluation}

A second key challenge is that ADS often intervene in organizations and institutional environments with complex interdependencies, enhancing possibilities for unintended or unanticipated spillover effects \citep{laufer_optimizations_2023}. Although an ADS may initially be designed for a well-bounded decision task within a specific subunit of an organization, predictive information can travel beyond its original purpose and shape broader institutional or system-level dynamics. Predictions developed for one kind of decision may inform others, such as when credit scores inform hiring or when pretrial risk assessments influence plea bargaining dynamics \citep{rona-tas2017offlabel}. 

These spillover effects raise thorny questions: Where and how should we bound the evaluation of ADS? Should assessments be limited to the intended function of the system, or should they also account for downstream and second-order effects? Tracking these broader adaptations may enhance understanding of an ADS's full impact, but it also complicates the task of evaluation. A robust evaluation framework must be flexible enough to detect important emergent effects without becoming impractical and unmanageably broad. This requires a combination of methodological approaches that can capture both intended and unintended consequences. In later sections, we outline possible strategies for achieving such a balance.

\subsubsection{Measurement Challenges and Goodhart's Law}

A third critical challenge in evaluating ADS is that measurement itself can influence or distort the behaviors and processes being assessed. This dynamic is captured by Goodhart's law: ``when a measure becomes a target, it ceases to be a good measure" \citep{goodhart1984problems}. Once actors -- whether individuals, organizations, or entire institutional systems -- know they are being assessed on a particular metric, they may optimize for that metric in ways that distort its intended purpose.  Scholars from across disciplines have documented the challenge of differentiating between genuine improvements in performance and strategic gaming of metrics without actual progress on the underlying dimension being measured  \citep{hardt2016strategic,kleinberg18investeffort,laufer2023strategic}. This issue is particularly pronounced when evaluative measures intersect with professional incentives or performance measurement systems. Qualitative evidence from \citet{saxena2021admaps} also highlights that algorithms used in the child welfare risk assessment ecosystem often lead to ``gaming'' behaviors, such as caseworkers manipulating the inputs to achieve the desired outcome or manage conflict, as that is the only way for them to regain agency within a system that limits their discretion.\footnote{Similarly, \citet{nader2019} documents that the Board of Veterans Appeals lowered the standard of review to be able to report consistently high ``accuracy'' of adjudicated cases, posing a sharp tension between the goals of error correction and performance reporting.} 

Aside from gaming, measurement asymmetry further complicates ADS evaluation.  Quantitative measures tend to attract more attention and command greater authority in organizational contexts, and they may thus drive resource allocation and decision-making because they can be more readily tracked and compared \citep{porter_trust_1995, espeland_sociology_2008, chu_cameras_2021, chang_does_2024}. This can lead to a bias in which what gets measured quantitatively is elevated and targeted for action \citep{muller_tyranny_2018}, regardless of its relative importance to stakeholder values or decision quality. Considerations that resist easy quantification, such as dignity or autonomy of decision subjects, may be systematically undervalued in evaluation frameworks despite potentially being central to the social contexts in which ADS operate \citep{yurrita_towards_2022}. This asymmetry creates a misalignment where evaluations privilege what can be counted over what individuals or communities believe should count. Such blind spots can produce evaluations that paint a rosy view of impact on quantitative metrics while obscuring an ADS's failure to deliver on broader institutional goals and values. This challenge is evident in hospital ranking systems based on average treated post-hospital outcomes, which can create misaligned incentives where hospitals may avoid high-risk patients to improve their rankings rather than focus on improving patient care \citep{wang2024counterfactual, dranove_is_2003}. 

Finally, the evaluation infrastructure itself may influence behavior. In more active research designs like randomized controlled trials, participants' awareness of being evaluated can shape their degree of engagement with an ADS. For instance, healthcare providers in a trial evaluating a diagnostic support tool might be more attentive to its recommendations than they would be during routine clinical practice, raising questions about the external validity of such evaluations. Randomizing provision of ADS outputs across cases (for example, as in \cite{ben2024does}), moreover, may increase the relative salience of the information provided. The observed responses might thus not always reflect what we would observe in the real world outside of the context of a formal evaluation.

\section{Implementation Science}
\label{sec:implement}

How do we know if an ADS intervention is implemented well in practice? In many cases, the differences in deployment effectiveness boils down to context - under which assumptions or expectations and in what kind of environment was the system released? Capturing and characterizing this context across different scales is central to \emph{science} of ADS implementation.
We define \emph{\textbf{implementation science}} as the \textbf{systematic study of methods and strategies that support the integration and sustained use of evidence-based ADS in real-world settings}, drawing from healthcare and education literature \citep{bauer2015introduction,kelly2012handbook,moir2018implementation, wensing2015implementation}. 

In this section, we describe the challenge of ``implementation design", which integrates context
- ranging from the narrow context of human-AI interactions to a broader institutional and societal context - in deployment practices of machine learning systems. 

We explore what defines the context in which an ADS system is deployed and how this can actively shape system outcomes. Anything from capacity constraints to the underlying regulatory environment to inherent human decision-maker biases are factors that ultimately impact deployment outcomes in meaningful ways without involving any kind of update to the ADS system directly -- instead, this impact is mediated through choices that define the bureaucratic context in which the system is released. We thus ask -- what are the environmental and contextual factors of model deployment that most directly impact model effectiveness? 

Unlike evaluation science and even model design, which define and characterize the examined model, implementation science is pre-occupied with defining the operation of this model ~\emph{in situ}, and characterizing the context that defines the outcomes of the broader system or policy into which the ADS system is being introduced. 

\begin{boxD}
    \paragraph{Key Takeaways} Understanding ADS deployment requires a clear understanding of the context in which the system is deployed - this context can range from the hyper-local considerations of individual user decision-making to the broader institutional and societal environment into which the ADS system is released. Quantitative and qualitative measurements of various deployment factors can inform our understanding of the differences in expectation and performance between otherwise identical models, implemented under varying contextual realities. 
\end{boxD}

\subsection{Individual Deployment Context}

First, we consider the context of the individual decision-maker. In behavioral economics, it is well understood that the choices and preferences of individuals can change, and these changes are largely influenced by the decision-making context~\citep{kahneman2003maps}. Here, we consider two important ways that deployment context impacts behavior at an individual level.~\emph{Interaction design} relates to the context in which individuals consider information and make the decision at-hand - this includes the ways in which a prediction score might be presented or explained to the decision-maker, the time available to make the decision and so on. ~\emph{Intervention design} at the individual decision-making level relates to the context of how an individual stakeholder is meant to respond to a decision made by them or some other actor - this includes the availability of certain actions in response to the prediction and street-level discretion of the individual in how they not just interpret and interact with the ADS system, but also how they respond to the prediction.

\subsubsection{Interaction Design}

Humans interact with predictions with different degrees of agency and awareness.

\begin{figure}
    \centering
\includegraphics[width=0.6\textwidth]{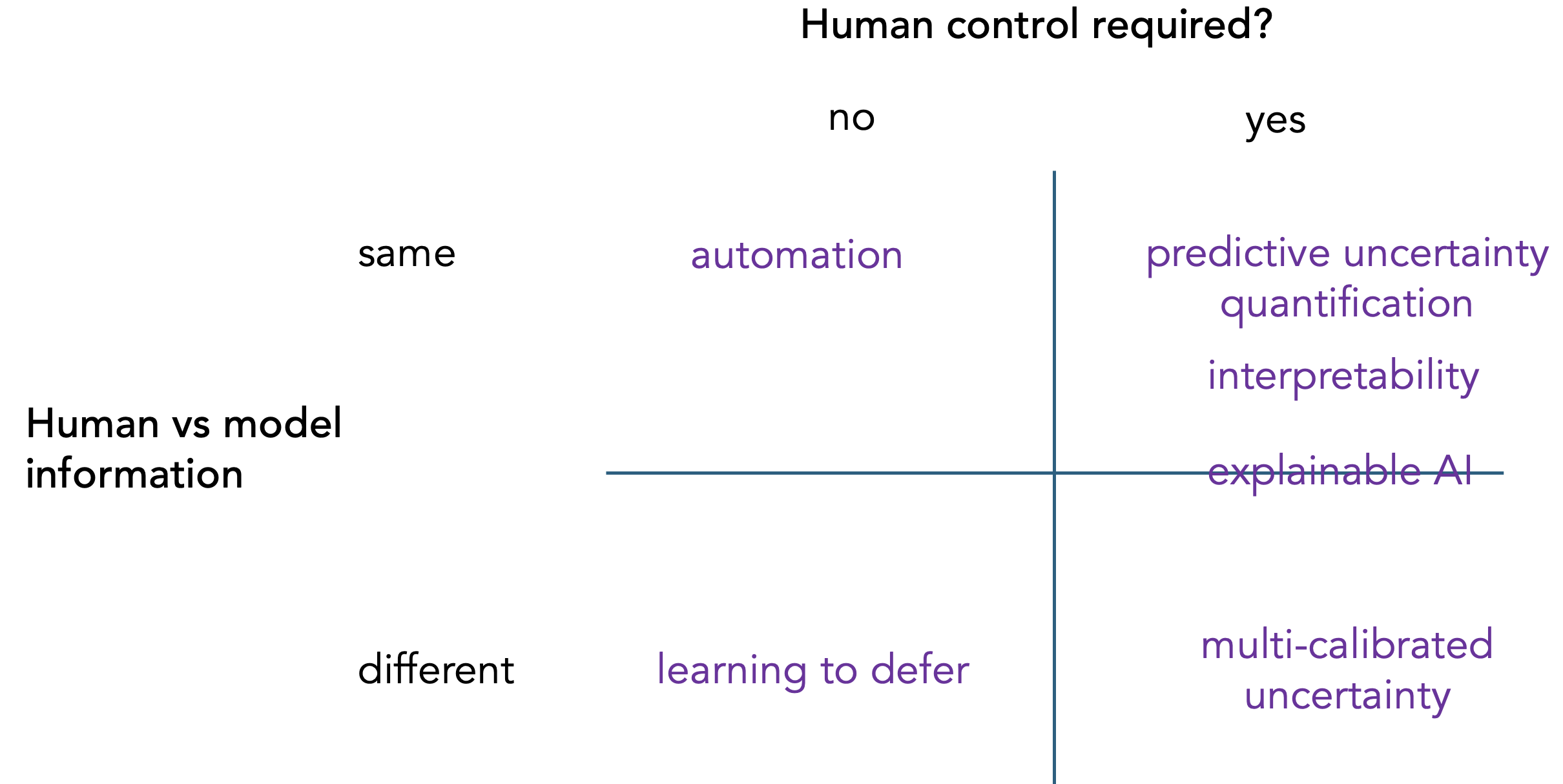}
    \caption{}
    \label{fig:hci-quadrant}
\end{figure}

Many predictive models are deployed in decision pipelines where humans have historically made decisions. 
Clarifying specific details of the problem formulation in institutional context illuminates how best to integrate a statistical model into a decision context. 
A first question that arises is, ``Must human control be maintained over the final decision?", reflecting constraints on the institutional decision-maker. A second question is, ``Is the human likely to have additional or different information than what the model was trained on?", which requires knowledge of current processes. Crossing these questions results in four distinct scenarios, shown in \Cref{fig:hci-quadrant}, which implicate different implementation strategies.

The question of whether we should expect a statistical model or a human with access to the same information (left column of \Cref{fig:hci-quadrant}) to perform better has been well-studied under the guise of clinical versus statistical prediction~\citep{meehl1954clinical}. 
Empirical studies have found that human decisions underperform those of statistical models (on average over the same distribution as used for learning); both earlier works prior to the advent of modern deep learning \citep{aegisdottir2006meta,grove2000clinical}, as well as other recent studies afterwards \citep{bansal2021does, buccinca2020proxy}. 
Consequently, when the human and the model are expected to have roughly equivalent information, and a human is not required to make a final decision, automation is likely to perform better (\Cref{fig:hci-quadrant}, quadrant 1).

In other scenarios, the two types of agents are expected to have similar information, but a human must remain in control of the final decision for reasons of liability or accountability (quadrant 2). In medicine, for example, automating decisions about treatments may be considered out of the question. The implementation challenge thus becomes how to persuade a human decision-maker to accept the model’s advice as often as possible. A typical assumption is that people are more likely to take advice when they can understand about how it is generated. This motivates using interpretable models rather than blackbox models; i.e. algorithmic transparency can improve compliance. Such interpretability can take the form of constraints in form, restrictions in size, or alignment with domain knowledge \citep[see e.g.,][]{ustun2015slim,ustun2019riskslim}. 

Providing information on the model’s accuracy and training the user on examples where ground truth is provided can help encourage trust \citep{mozannar2022teaching,thompson2021mental}. Uncertainty information is also useful \citep{bhatt2021uncertainty}, such as the uncertainty or confidence associated with each recommendation. This all can inform the best (i.e., under rational assumptions) action that can be taken by the decision-maker when the model encodes all relevant known information \citep{hullman2025underspecified}. 
If we assume that decision-makers have internal models of the relationship between instance features and states of interest but are boundedly-rational in the sense of facing cognitive constraints, explanations of how a model arrives at a prediction (i.e., functions applied to instance features $X$ and model predictions $\hat{R}$ or $\hat{Y}$) may help them estimate the probability that a prediction is correct through comparison to their own knowledge \cite{hullman2025explanations}. 
					
In some settings the human and the model have access to potentially differing information, and human control over the final decision is not required (quadrant 3). Here, the question becomes how to best distribute test instances to the two types of agents. One approach is rejection learning \citep{cortes2016learning} which puts aside instances with high predictive uncertainty. However, better performance can be gained by a learning to defer framework that jointly learns a predictive model and a rejector that determines whether the classifier or the human should predict \citep{mozannar2020consistent, mozannar2023should}, as this approach takes into account the human’s expected performance as well.

When the human and the model have access to potentially differing information, but the human is expected to retain final control over the decision (quadrant 4), the challenge becomes how to best inform the human decision-maker about their expected performance versus that of the prediction model. 
Using feedback from realized outcomes to extract
information on human versus model performance over the feature space could inform probabilistic predictions
on the current instance. 
Methods for quantifying prediction uncertainty, such as conformal prediction or posthoc scaling of model-estimated confidence like softmax, must be applied with care when the human may have access to unique information \cite{hullman2025conformal}. 
Calibration of methods 
should take into account both agents’ chances of being correct to prevent misleading the decision-maker, for example with multi-calibrated uncertainty \citep{corvelo2024human} over
the model’s information as well as the human’s confidence function. 

Additionally, although human decision-makers each have default biases that impact how much they pay attention to an ADS system, there are also operational context factors that may impact individual human judge responsiveness. For example, a consistency bias in human judges can make it so that if the prediction is only shown intermittently for cases, then the information provided by the ADS is more likely to be ignored, compared to instances where this information is consistently present - this has implications on, for example, the ideal treatment assignment in experimental evaluations of ADS, with a case-based randomization scheme having a much less notable causal effect than decision-maker level randomization~\cite{raji2025evaluating}. 

\subsubsection{Intervention Design}

Another critical part of individual context is the specific \textit{intervention design} of what would be done in response to ADS information. Institutions must also consider designing interventions for effective ADS integrations.


For example, improving outcomes in education with ADS  requires not only simply identifying at-risk students, but also thoughtful, evidence-based interventional strategies.  \citet{liu2023reimagining} challenged the assumption that surfacing risk scores is inherently useful, arguing  that `naive' interventions such as showing risk scores to students and schools without providing additional resources and support can hurt more than help.
\citet{mcconvey2023human} found that interventions in response to ADS in higher education took various forms, including automated email notifications, supplemental coursework, and advising meetings, but none showed consistent empirical evidence of improving student outcomes. They highlight the potential of iterative, participatory design approaches that involve learning strategists and other stakeholders in developing theoretically grounded interventions alongside ADS model design. 

Recent work by \citet{Schechtman2025discretionloophumanexpertise} also suggest that, even when algorithmic decision support is available, human discretion (e.g. by a college advisor) still play a crucial role in personalizing educational interventions to each student. Effective intervention design should account for where and how human participants (i.e., by teachers, counselors, and advisors) add value within the decision-making loop.

\subsection{Institutional Deployment Context}

The implementation of an ADS system is far from simple. The field tends to underestimate the complexity of the institutional context in which ADS systems are integrated and deployed.  Default institutional decision-making processes (i.e. the ``bureaucratic counter-factual'' to an ADS deployment) already tend to be complex, typically involving multi-step procedures executed by a variety of stakeholders, each with their own intertwining incentives and constraints~\citep{klein2017sources}. 

\begin{figure}
    \centering
\includegraphics[width=0.7\textwidth]{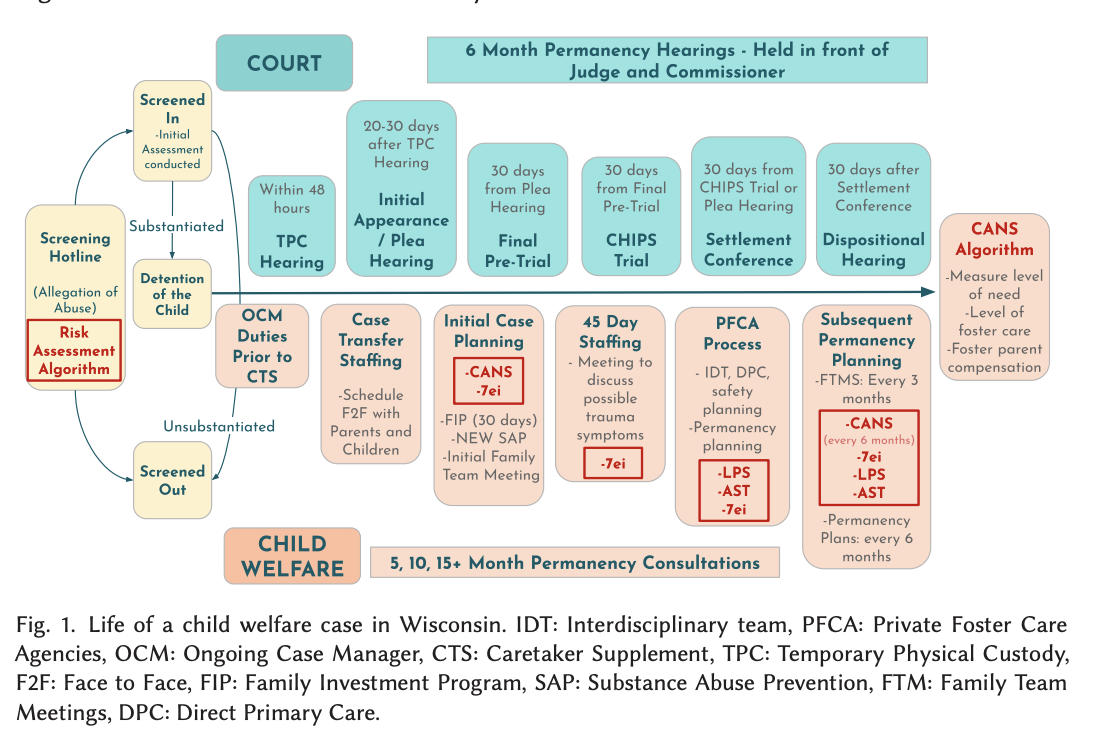}
    \caption{}
    \label{fig:child-welfare}
\end{figure}

Beyond the ``bureaucratic counter-factual'' as a baseline institutional policy, an ADS deployment is often even more complex in practice. 
ADS integrationss are an institutional policy change involving \textit{multiple} steps and \textit{multiple} stakeholders, \textit{each} of which can, in some way, impact institutional outcomes (See Figure ~\ref{fig:child-welfare} for such a journey map illustrating this complexity). 

\subsubsection{Integration \& Change Management}

Integrating and updating a prediction-based intervention into existing organizational policies and workflows can be an incredibly complex process. Typically, this begins with developing buy-in from high-level institutional stakeholders to introduce the ADS system into the organization in the first place, and can involve negotiations with a variety of internal stakeholders to participate in model development, training and integration. 

As an example, the ``journey maps'' for the implementation of the Duke Sepsis Watch ADS system are incredibly complex ~\citep{boag2024algorithm} and illustrates the wide variety of involved stakeholders.
The journey map spans from the executive hospital management who approve funding, to the nurse practitioners who develop and train on-the-ground users, to the regulatory compliance teams who vet the model's suitability for deployment given legal and institutional restrictions. Well-implemented prediction-based interventions need a detailed change management plan to ensure meeting new expectations and addressing unexpected challenges. 

\paragraph{Bureaucratic Counterfactuals \& Policy Decision-Making Pipelines} 
In reality, the bureaucratic counterfactual we have been referring to in abstract comprises multiple process steps and stakeholders. Interventions made \emph{without} predictions still require a sequence of decisions, each of which informs the probability of another set of choices. The judges, $J_{k,i}$ making such decisions, $D_i$ are at times distinct from the actors $A_{l,i}$ acting on an intervention $I_i$. These actors are often distinct from the decision subject $S_m,i$ and the creator (procurer or developer) of the model prediction $C_j,i$. Fully mapping out the bureaucratic counterfactual with or without ADS can be complex.

Further, 
most deployments manifest as multi-prediction pipelines, not single model predictions at a single decision. They require multiple models used at different stages of a complex decision-making process, interacting with distinct stakeholders and other models as part of a systematic ADS pipeline~\citep{saxena2024algorithmic}.

\subsubsection{Implementation Factors \& Considerations}



Considering a simple question highlights the complexity of how the deployment context can shape system outcomes:
``What is the difference between predicting risk in (A) criminal recidivism and (B) hospital re-admission?''
These prediction tasks (as well as similar tasks such as e.g. child-welfare alert algorithms~(\cite{saxena2023}), have an almost identical problem formulation -- input characteristics of the model subject predict the probability of a negative outcome (i.e. the inappropriate release of the model subject, requiring re-entry into the institutional system) and thus inform interventions to minimize that negative outcome in practice. Although the prediction task of clinical re-admission and criminal recidivism seem on their face to have similar properties (ie. how likely is the release/discharge to fail and the model subject to return?), several conditions of deployment and integration practices can lead the prediction models to have vastly different degrees of impact in-situ. 

Moreover, even within specific domains, prediction tasks that sound like similar prediction tasks yield vastly different decision-making processes. For instance, within child welfare systems, predicting risk of re-referral and risk of discharge often use similar data inputs and modeling strategies yet, qualitatively mean very different things to the decision-makers implementing interventions in-situ (\citep{saxena2023}).

In order to explore the range of ways in which context can factor into prediction outcomes, we look at the impact of context from the most narrow (single human-AI interaction) to more broad (eg. institutional and societal level context). 

\begin{figure}
    \centering
\includegraphics[width=0.5\textwidth]{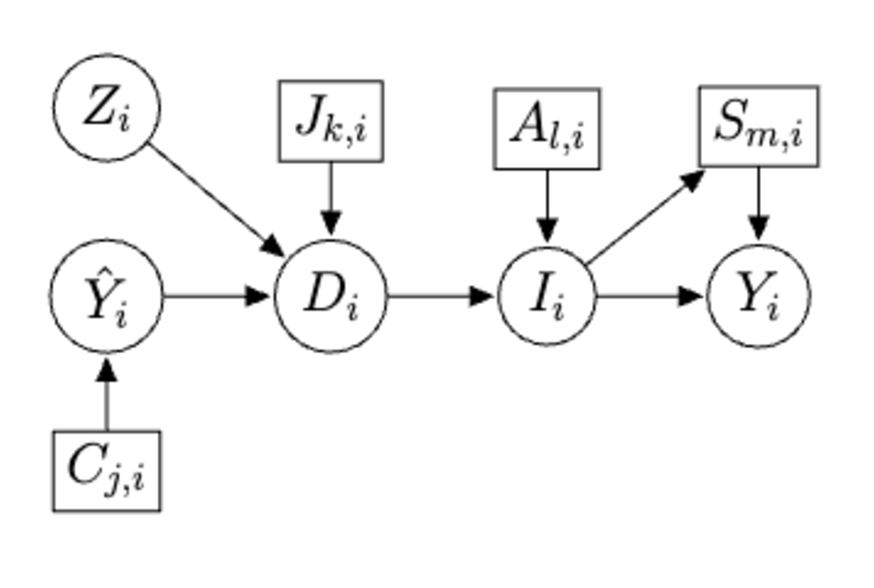}
    \caption{Diagram illustrating the many stakeholders involved in administering and making decisions about individual cases. 
    }
    \label{fig:Imp_Diagram}
\end{figure}

We 
map out the involved stakeholders and key state variables in Figure ~\ref{fig:Imp_Diagram}, 
to illustrate the differences between the scenarios that go beyond the scope of the typical modeling considerations. 

What are the stakeholders that impact the outcome? \footnote{Note that each of these stakeholders can preside over multiple cases $i$.For instance, $S_{m,i}$ can be a decision subject over time across multiple cases $i$.}
\begin{itemize}
    \item Creator of the algorithm , $C_{j,i}$, that introduces the prediction model into the system, either through procurement or model development. 
    \item Judge, $J_{k,i}$, which makes the decision $D_i$
    \item Actor, $A_{l,i}$, which enacts the intervention $I_i$
    \item decision-subject $S_{m,i}$, (which can possibly impact future prediction through performativity or other feedback loops?)
\end{itemize}

\paragraph{Stakeholder Incentives}

Each deployment scenario has its own distinct set of stakeholders -- each of these stakeholders has their own incentives and priorities, which may or may not align with that of the decision subject \citep{laufer2023strategic}. 

For instance, for task (A), the judge (i.e. the decision-maker), is attempting to balance public safety concerns with the well-being of the criminal defendant (i.e. decision-subject). In such a case, the preferences and priorities of the judge and the defendant are not fully aligned, and there are actions the judge can take that will be directly harmful to the defendant, but aligned with a favorable outcome for them as a judge. There are even differences in alignment between the impacts of decisions for society at large (social costs of detention or crime) vs. the judge themselves, as cases where judges release defendants who then go on to commit other crimes often receive inordinate attention and potential blowback for judges. \citep{klllm17} refers to this as potential ``omitted payoff bias" arising from this misalignment. However, for task (B), the doctor and the patient are fairly aligned stakeholders -- insurance regimes such as US Medicare will not pay both the patient and healthcare provider if re-admission rate is high. Beyond insurance considerations, re-admitted patients introduce congestion costs and additional load on system, while of course patients want issues resolved without re-admission. Not only do individual decision-makers face these incentives, but also the institutions that decided to introduce ADS in the first place.


\paragraph{Timing}
The timing of display and the presentation format of the prediction has an impact. For task A, the prediction is shown as a stratified band (high-mid-low risk) to the judges \emph{before} the decision is made to release or retain the defendant.\footnote{Presentation format matters, though this can be difficult to formalize. For example, risk scores summarize prior criminal history with flat quantitative information, which might otherwise be presented physically via printed reports on each prior encounter. The latter format can appear more salient. The PSA decision-making matrix used to be circulated with ``red" emphasis on higher-risk designations vs lower-risk designations, even though absolute risk of high PSA scores were low overall. It has since been updated to use varying shades of green instead. 
} For task B, the re-admission score is shown after the discharge of the patient, in order to inform which resources are provided to the patient following release.

\paragraph{Decision Stakes}
Both prediction tasks A and B can be used as part of quality control for the decision-maker -- i.e. a hospital's funding and quality rating can be contingent on the number of hospital re-admissions they allow, and a judge's performance could hypothetically be evaluated by considering the recidivism rate of assigned defendants.

However, the stakes for the decision-subject vary. For task A, a score close to threshold could lead to jail time and a meaningful lack of freedom while awaiting trial, at times for years. Meanwhile, the hospital re-admission score for an individual patient only results in whether or not they are nudged towards certain additional resources post-care. There are potential secondary effects from the resource allocation differences caused by this prediction, but ultimately the stakes of the intervention and the decision of the subject of the decision are much lower. 

\paragraph{Capacity Constraints}

Capacity constraint differences can make it difficult to operate in the context of limited available resources. This impacts everything from user responsiveness to intervention effectiveness. 
Once operating at a particular capacity, especially in high stakes settings, decision-makers become less responsive to input information and nudges, as they are no longer in a position to act on the algorithmic recommendation past a certain threshold of capacity to respond. This is a finding that has been observed pragmatically in many settings by decision theory scholars ~\citep{klein2017sources}, and economists ~\citep{boutilier2024randomized}. The evidence is somewhat mixed however on the broader implications for generally resource constrained settings, where it has been observed by computer scientists that decision-makers may opt in those cases to over-rely on algorithmic recommendations, rather than invest in more complex, independent decision-making ~\citep{buccinca2021trust, goddard2012automation}. For example, human decision-makers who were previously accustomed to using their judgment to resolve ambiguities in the framing of a decision problem, such as which information is relevant to consider (e.g., a doctor using their discretion to decide whether to consider still preliminary evidence on previously overlooked side-effects when prescribing a drug), may be less inclined to constructively reason about such questions when supplied with an ADS~\citep{felin2025artificial}.  
This is still an area of ongoing discussion and study.

\subsection{Societal Context}

\paragraph{Application Domain}
Differences or similarities in institutional contexts across scenarios give rise to key contrasts or similarities in implementation design such as the legal ecosystem, the scale of the decision-making, the stakes, recourse or reversibility of a wrong decision or action, among others. 
For example, the medical and financial contexts are much more regulated than, for example, criminal justice or education deployment contexts. Judicial criminal sentencing guidelines have substantial room for inherent discretion, compared to medical clinical decision-making guidelines for diagnosis or prescription.

\paragraph{Regulatory Compliance \& Legal Compatibility}


Prediction-mediated decision-making falls under the same legal and regulatory jurisdiction as the deploying institution, and is often 
prone to similar if not more legal scrutiny than typical decision-making. In fact, the legal compatibility of evaluation and implementation is necessary for a predictor to be operable in deployment and legally valid~\cite{xiang2019legal, raghavan2024limitations}.

\section{Prescriptive Angles 
}\label{section-research}

The prior sections outline a more expansive view on the design and evaluation of ADS, arguing that a benchmarking predictive accuracy rarely provides the evidence needed to warrant deployment or improve  them. In what follows, we share a few different \textit{prescriptive angles} that describe other potential paths to move forward and expand current practices. 

\subsection{Path 1: 
Interrogating the Role of Prediction}

\begin{boxD} \textbf{This path forward involves interrogating the role of prediction.} Some may advocate for replacing prediction-based algorithms with alternative decision processes that address systemic injustices directly.
\end{boxD}

Our prior analysis implies that prediction models are helpful inasmuch as they inform better decisions. In medicine, new prediction models are published daily – yet some critical appraisals question their clinical utility and point out that very few are used in clinical practice~\citep{wynants2020prediction, van2022critical}. Prediction by itself doesn't necessarily bring about better outcomes: it's necessarily to articulate mechanisms for how they would do so, walking through how prediction information should change downstream decisions and outcomes..

Overall, we have focused on socio-technical questions regarding evaluating ADS. 
But our analysis of institutional context underscores that these tools are integrated in a narrow scope of existing processes, and are thus limited in their capacity to improve them broadly. Technical concerns re: developing, evaluating, and improving ADS are necessarily just a small part of broader discussions of social change and improvement. (See \citet{abebe2020roles} for discussion about how computing research may nonetheless have important roles to play in social change - but rarely from discussing technical prediction aspects alone.)

It's important to be consider expansive alternatives beyond prediction: advocating for more resources, improving intervention effectiveness, and many others. 
One example alternative is that of \textit{lotteries} (including partial, weighted, or tiered lotteries), where interventions are randomly assigned among a possibly pre-screened pool of eligible individuals, pose an alternative approach. 
Lotteries could provide a fairer \citep{stone2007lotteries} and more transparent mechanism for resource allocation, as demonstrated in healthcare \citep{white2020proposed}, housing \citep{arnosti2020design}, and research funding \citep{horbach2022partial}.  

In the end, it is important to critically evaluate whether algorithmic development is necessary; sometimes the best solution is not an algorithmic one.

\subsection{Path 2: Better and More Engineering}

\begin{boxD}
\textbf{This path forward seeks creative, thoughtful, and appropriate engineering improvements that are better aligned with improving social outcomes.} While researchers might disagree on exactly what these are, these can include
changing the estimand from predictive to interventional, refining predictive models and data, or enforcing additional desiderata (e.g. fairness, better target data).
\end{boxD}

\textbf{Changing the estimand:} When algorithmic solutions are desirable, the appropriate tools may not always be \textit{prediction} models, but interventional or decision-aware approaches. \Cref{sec:model_design} outlines these alternatives in detail, including their technical formulations. Changing the problem formulation, changing the estimand, or introducing additional desiderata can all be crucial tools for better aligning ADS with outcomes of interest. Alternative approaches include identifying causal determinants as candidates for intervention, optimizing treatment rules, or estimating counterfactual risk models. 
 
Some scientific tasks are not most fruitfully framed as prediction tasks. 
It is important to distinguish causal from predictive models because of fundamentally different technical considerations that arise, differences in relevant criteria of success, and differences in the kind of information available or necessary to appropriately evaluate causal models.

\textbf{Revisiting data collection} Researchers working on ADS systems often treat the data available as fixed, relying on existing types of administrative records to train predictive models. 
Yet there may be many cases where researchers should consider whether it is feasible to collect or access additional data, and what the costs and benefits of such a process would involve. When the true outcome of interest is not collected in existing data,
 it is worth evaluating whether it is feasible to do so, what are the potential benefits, and costs. Indeed, the social cost of tracking individual trajectories can be extreme. Trading off these costs might suggest triangulation based on population statistics.  



\subsection{Path 3: Understanding and assessing models in deployment}\label{path:holisticdeployment}
\begin{boxD}
\textbf{This path forward is to develop holistic frameworks for evaluating models in deployment, combining rigorous experimental design, participatory approaches, and qualitative assessments to ensure appropriate use and address diverse harms.}\end{boxD}

While we have focused on welfarist accounts of the utility of algorithms, understanding their deployment and procedural transparency/opacity is crucial. Algorithms are often painted in context as either opaque black boxes \citep{pasquale2015black} or more transparent alternatives to the black-box of \textit{human} decisions \citep{mullainathan2019biased}, depending on context and use . 
In many consequential domains, procedural due process protections are required. This can take the form of appeals processes or human review, which can be important for contesting incorrect algorithmic inputs.

The affordances and requirements of algorithms change significantly depending on context. An algorithm can simultaneously make some aspects of a decision more transparent (e.g., the ordering of inferred risks across individuals) while making others more opaque (e.g., the process through which risk scores are inferred). 
Algorithmic systems need to be vetted for both efficacy, safety, and procedural effects, especially in light of complex processes already in place for performing similar functions \citep{laufer2023algorithmic}.

This effort should consider the perspective of decision subjects in affected communities from conception. The algorithmic equity toolkit of \citet{katell2019algorithmic} facilitates community-driven audits of technology.\footnote{On the other hand, recent interest in incorporating participatory methods into AI systems has also spawned criticism of potential ``participation-washing", leveraging the presence of community input to justify extractive AI/ML systems without longer-term engagement or meaningful engagement with community needs. Developing meaningful co-design processes and making this standard process would be important.} Chicago Beyond, a social impact investor, developed a guidebook for community organizations, researchers, and funders to address inequitable power dynamics in social research \citep{chicagobeyond}. Their title comes from a question from a participant who asks, ``why am I always being researched?", as his peers are in research studies, and his family and even the staff at nonprofits administering programs recall being the subjects of research. This anecdote highlights how ``problem formulations" that exist in isolation for researchers exist in confluence for affected communities, especially for marginalized households navigating precarious realms of health, housing, labor, education, justice, and so on.\footnote{Another example of this disconnect occurs when algorithmic researchers study a problem formulation in abstract, and are surprised by the depth of community resistance or criticism, especially in consequential domains of surveillance and sanction. While researchers may discuss such systems in terms of an abstract ideal, and give institutions the benefit of the doubt from a neutral position, communities are ultimately the one subject to potential harms. They are more concerned about the tendencies of institutions as they are. }

\subsection{Path 4: Sustainable Governance and Maintenance of ADS}

\begin{boxD}
\textbf{This path forward is to establish sustainable governance and maintenance for ADS by institutionalizing accountability, assessing costs versus benefits, codifying practices, and fostering policy engagement via regulatory conversations.}
\end{boxD}

So far, our analysis has heavily emphasized the nuances of evaluating the impacts of ADS. This implicitly assumes the capacity and willpower to evaluate the impacts of ADS at all, whether prospectively or retrospectively. In many settings, perhaps due to vendor secrecy or procurement environments, this is not possible --- potentially at great harm to individuals. 

This has happened repeatedly in the case of the automation of state benefits. 
One example is that of the lawsuit K.W. v. Armstrong, a class-action lawsuit brought by the ACLU representing disabled Idaho residents receiving assistance from the state's Medicaid program \citep{acluidaho}. Benefits were being cut 20-30\%, with no explanation or recourse. Over extended class-action litigation, the ACLU reverse-engineered the system with experts and found many essentially arbitrary determinations.\footnote{There were two assessments, one developed in-house without any validation,
standardization, or auditing; the other developed by a private company. Both the State and the company fought to prevent access to the assessment booklet (instructions for assessors to compute scores), under the guise of trade secrets. But access was necessary for due process and to challenge errors in their  determinations.} Although the judge ruled that the assessment must be disclosed and tested, litigation continues to ensure the state follows through. 

Eppink, an attorney Of Counsel, concludes in a testimony before the Senate on ADS governance, ``The lesson is this: Decades-long class actions by indigent families are not a viable
plan for AI governance in federal programs." Instead, auditing and performance evaluation and management of ADS ought to be monitored and regulated as part of regular procedure.

Developing mechanisms and governance structures for performance management is necessary when it is not already present. It's still impossible to anticipate all harms ahead of time, so experimenting with new structures for continual performance monitoring is crucial. Although incident report databases such as those used by vaccines~\citep{vaers} have some issues with implementation (e.g., selective reporting, lack of knowledge by users), they can provide a feedback mechanism for deployers. 

\paragraph{Acknowledgments}
We are grateful to all workshop participants for discussions.

\clearpage 
\appendix

\textbf{Appendix}

\addtocontents{toc}{\protect\setcounter{tocdepth}{-1}}

\section{Additional discussion on Background}

\subsection{Additional examples}\label{app:examples}

\begin{example}[Criminal justice - Further Policy Context for \Cref{ex:justice}]
    New Jersey's use of risk assessment tools is widely viewed as a success in terms of reducing jail populations (although racial composition of the jail population did not change, dispelling claims of RAI proponents that RAIs operate with ``less racial bias" than judges and therefore might reduce racial disparities) \citep{aclunjJerseyUsed}. However, its use of the PSA accompanied earlier commitments to abolish money bail and was generally part of a larger raft of structural reforms \citep{anderson2019evaluation}. This question of policy regime and alternatives is essential: California's Proposition 25 would have replaced money bail with pretrial risk assessments, without specific protections on the performance or racial bias of such tools, among other concerns. Conversely, new policy mandates declared from executive offices need not be executed if there is wide disagreement among the ranks, i.e. among ``street-level bureaucrats" who make final decisions. For this reason, \citep{cowgill2020algorithmic} argues that attempts to embed particular social preferences within algorithmic predictions could backfire if such preferences conflict with final decision-makers, who might ignore such predictions  outright. 
\end{example}

\begin{example} 
[Wisconsin Public Schools Dropout Prediction]
Dropout Early Warning System (DEWS) \citep{Knowles2015DEWS} was a risk scoring system
designed and implemented by the Wisconsin Department of Public Instruction (DPI) from 2012 to 2023 \citep{WisconsinDEWS}. The system provides educators with risk scores for all middle school students (grades 6-9), generating these predictions using over 40 student features.  Features include demographic and socioeconomic information (e.g., race, gender, family income),
academic performance (e.g., scores on state standardized exams), as well as community-level statistics
(e.g., percent of cohort that is non-White and school size).  The system has two main outputs: 1) the DEWS
risk category (low, moderate or high) and 2) the DEWS score (an estimated
probability of on-time graduation that takes continuous values between 0 and 1). Models are fit by empirical risk minimization on a dataset of a historical
cohort of students, and are updated every year. While DPI maintains that DEWS dropout predictions ``can lead to critical interventions that prevent students from actually dropping out.'', guidelines provided to schools on how to act on the DEWS scores leave out key model information and provide insufficient guidance for interpreting these scores \citep{feathers2023takeaways}. There has been no systematic data collection on the actual educational interventions applied in response to DEWS.
\end{example}

\begin{example}[Health - Liver transplantation]
    For example, consider a clinical decision support task, deciding which patients with liver disease are good candidates for liver transplantation. One approach would be to predict the expected survival time under a scenario of no transplant and prioritize patients (in part) based on this counterfactual prediction, those who die soonest without transplant. Another approach would be to predict expected survival times under two or more scenarios: e.g., no transplant, immediate transplant, or transplant after some delay. 
Depending on the data and context, researchers may be better positioned to predict one or several of these counterfactual scenarios accurately (or relative differences between them, which would be used in an ITR). 
\end{example}
\begin{example}[Health - diabetes prediction; how to choose predictors]
    An example of these questions arise in the case of diabetes risk prediction. Existing work has shown that there is a difference between diagnoses for diabetes from a physician vs. blood tests \citep{coots2025framework}. Knowing this difference exists, recent research has demonstrated that including data on health insurance degrades the predictive performance of the model on actual diabetes \citep{mikhaeil2024hierarchical}. Here, researchers should consider the benefits vs. costs of creating training data with true diabetes test results, and potentially developing ongoing performance monitoring processes. 
\end{example}

\section{Additional discussion on Model Design}

\subsubsection{More detail on WPRS}\label{sec:model-design-apx-wprs}

Local differences in capacity lead to a “tie-breaking” design where local offices provided mandatory services for highest-risk claimants, randomizing at the discrete score value that satisfies the capacity constraint. The use of mandatory services was thought to improve moral hazard. While the treatment group overall returned earlier to work, improving total earnings, the timing of these early returns coincided with the onset of mandatory work requirements. 
The variability in local offices allowed for assessing treatment effects in quantiles of risk, finding evidence against a ``common effect” assumption that justifies prioritization by expected duration (in the absence of the program). That is, $E[Y(1)-Y(0) | Y(0)]$, as assessed via low, moderate, and high scores, was highly nonlinear in the predicted risk, with the greatest benefits accruing to those with moderate risk of exhausting benefits, overall providing “evidence that allocating the treatment on the basis of the expected duration of UI benefit receipt may not represent an optimal strategy.”

Trade-offs often arise when limited budgets are available for additional resources, and choices need to be made as to who receives them. \citep{berger2021evaluating} consider the varying equity and efficiency goals that might be relevant to decision-makers (``bureaucrats"). Their thought experiment is in their previous context of allocating services for reducing the duration of unemployment spells, and they ``suppose that our bureaucrat believes that the duration of a person's unemployment spell is a good indicator of their welfare: the longer the spell of unemployment the worse off is the recipient. Helping long-term recipients, which is desirable on equity grounds, may conflict with the objective of limiting the expenditures of the UI system, which may be desirable for political or efficiency reasons." Indeed, duration of unemployment spell (in the absence of services) might itself in general be positively or negatively correlated with the heterogeneous treatment effect, so that equity goals of improving the welfare of the worst-off might be perfectly aligned with or opposite \citep{berger2021evaluating}. 

\subsubsection{Technical formulations: Decision-aware learning}\label{sec:model-design-apx-decisionawarelearning}

\paragraph{Decision-aware learning.} 
Another class of approaches, variously called smart predict-then-optimize, task-based learning, decision-focused learning, and decision-aware learning, among other names, is typically motivated by complex action spaces, which often lead to a complex mapping $Y \to D$.  

\begin{parable}[Structured system-level decisions]
    Consider a problem with a combinatorial, constrained action structure like planning the shortest path in a graph. Suppose we wish to find the path with the minimum total travel time when the travel times on each of the edges is unknown in advance and must be predicted. 
\end{parable}

The decision rule approach would now have to directly map from features to a valid source-target path. Even ensuring a feasible output is nontrivial. Effectively, we may suspect that this decision rule is now being forced to reinvent the wheel: we already know how to solve shortest path problems with known costs, and it would be easier to make use of this knowledge instead of re-learning it. The plugin approach is able to leverage our existing algorithmic knowledge; we would first predict travel times and then plug the predictions into an optimization algorithm to compute the corresponding path. However,  the constraint structure of the problem can be  informative about which predictive models are better are worse. For example, errors that do not change the path selected have no impact on the loss, while even a small error can have large consequences if it changes the overall path. 

Decision-aware learning methods aim to modify the plugin approach to leverage this knowledge of problem structure. Specifically, they search for a predictor that minimizes the expected loss when plugged into the optimizer. Suppose that we optimize over a class of predictive models parameterized by $\theta$, denoted as $\hat{p}_\theta(Y|X)$. Decision-aware methods seek a model that induces the smallest possible loss when used in optimization:
\begin{align*}
    &D^*(X, \hat{p}) = \arg\min_{D \in \mathcal{D}} \int\ell(D, Y) \hat{p}(Y|X)dY\\
    &\theta_{\text{DA}} = \arg\min_{\theta } \frac{1}{n}\sum_{i = 1}^n \ell(D^*(X_i, \hat{p}_\theta), Y_i).
\end{align*}
The quantity that $\theta_{\text{DA}}$ minimizes can be called the ``decision loss" since it quantifies the loss when a model parameterized by $\theta$ is used for decision making. Finding $\theta$ that minimizes the decision loss is typically nontrivial. E.g., in our shortest path example, $D^*$ is defined by the output of a shortest path algorithm. A number of approaches have been proposed in the literature, which provide theoretically-grounded surrogates for the decision loss, attempt to ``differentiate through" the optimization in $D^*$, learn a parameterized approximation to decision loss, and more.

So far, we have assumed that the decision maker knows a specific loss $\ell$ that they wish to minimize. A recent line of work in machine learning seeks predictors which are near-optimal for decision making with respect to some \textit{class} of potential downstream losses.\footnote{
Technically, these formulations tend to be closely related to multicalibration, a solution concept which requires predictors to be conditionally calibrated for each of a (potentially large) number of subgroups. One example is the notion of an omnipredictor. Roughly, a omnipredictor relative to a set of loss functions $\mathcal{L}$ and a comparison class of models $\mathcal{C}$ guarantees that for every loss in $\mathcal{L}$, its predictions suffice to make decisions nearly as good as any model in $\mathcal{C}$. 
A related notion is decision calibration, introduced by \citep{zhao2021calibrating}. Decision calibration requires that for any loss $\ell$ in a set $\mathcal{L}$ and any alternative $\ell' \in \mathcal{L}$ the decision maker obtains lower loss (in expectation) by selecting $D \in \arg\min \E_{Y \sim \hat{p}(Y|X)}[\ell(D, Y)]$ than $D' \in \arg\min \E_{Y \sim \hat{p}(Y|X)}[\ell'(D, Y)]$. Intuitively, this can be understood as requiring that the decision maker cannot gain by ``post-processing" the predictions; they are incentivized to best-respond as if the predictions $\hat{p}$ are in fact correct. Similar ideas were recently generalized by \citep{noarov2023high} beyond the scalar case to predicting a multidimensional state followed by optimization. 
}

\subsubsection{Additional discussion on AFST}

{For example, the Allegheny Family Screening Tool (AFST) was reportedly under legal scrutiny under concerns about the use of disability status information. The Allegheny Family Screening Tool (AFST) is a predictive risk modeling tool that leverages Allegheny County's unusually rich linked administrative data, including information from Department of Health Services' ``integrated data system that links administrative data
from 21 sources including child protective services, publicly funded mental health and drug
and alcohol services, and bookings in the County jail" \citep{alleghenycountyanalytics}. Reportedly, attorneys from the DOJ consider potential civil rights concerns related to discrimination on the grounds of disability \citep{apnewsChildWelfare}; the AFST leverages information about presence in public programs (such as public benefits, SSI income) that is strongly correlated with disability. Some raise concerns that presence in these public programs is more of an indicator of poverty, and therefore the predictive risk tool's integrated information increases scrutiny and marginalization based on poverty (rather than potential supportive services) \citep{gerchick2023devil}. Although the tool's creators point out this might be justifiable on the grounds of increasing predictive accuracy, \citet{gerchick2023devil} argue this highlights embedded values in the design of the tool that might be contested under different perspectives. On the other hand, in evaluating placement recommendations from a new risk assessment instrument for homelessness, \citet{cheng2024algorithm} find that the current practice of leveraging an alternative risk assessment for new individuals without administrative history can introduce algorithmic bias, i.e. the additional administrative history improves predictions.
}
\subsubsection{Implications of embeddedness of decisions in the social world on problem formulations}\label{sec-apx-implicationsofsocialworldmodeldesign}

\paragraph{Modeling the ecosystem of predictive models and social dynamics}

While we have emphasized a wide variety of possible problem formulations focused on a single decision point, in reality predictive models are embedded in social dynamics. Default formulations focusing at a single decision point assume away potential dynamical issues. On the other hand, allowing unrestricted social dynamics is too difficult to predict and and it becomes difficult to draw any conclusions. 

Therefore, the key question is what kinds of social dynamics change what is best to do in the here-and-now. For example, strategic and adversarial responses illustrate the extremes, but best-response models can also be analytically tractable, and have been used to model the strategic ``gaming" of individuals in classification systems \citep{hardt2016strategic,perdomo2020performative}, ranking with strategic response \citep{liu2022strategic}, models of human capital investment and disparate equilibria \citep{liu2020disparate}, and longer-term impacts at equilibrium of short-term fairness interventions \citep{hu2018short}. 

In this vein, performative prediction \citep{perdomo2020performative} is a recent line of work \citep{mendler2022anticipating,miller2021outside,hardt2022performative,kim2023making,perdomo2025making} that explicitly aims to analyze the feedback loops between predictive algorithms and the social world. It is motivated by the fact that whenever we make predictions about social events, these predictions shape people's expectations and actions in ways that end up influencing the likelihood of the the predicted outcome and the distribution of observed features. For instance, predicting that someone is at risk of a heart attack influences their lifestyle choices in ways that shape their future outcomes.  Formally, performative prediction is a learning theoretic framework that, amongst other things, has been used to understand different kinds of social prediction dynamics such as repeated retraining. See \cite{pastandfuture} for an overview of work in this space.

However, interpretations of some theoretical results vary widely based on the relevant context, as well as the empirical validity of assumptions that were originally made for analytical tractability, rather than a mimetic accounting of social dynamics themselves.\footnote{For example, competitive dynamics converging towards a no-regret equilibrium is a better mimetic representation in discussions of algorithmic pricing in antitrust, because different firms do in fact run pricing algorithms that optimize based on customer demand and market conditions. On the other hand, in other situations, assumptions of equilibrium response or rational best-response are models that simplify out of necessity rather than mimetic fidelity to social dynamics themselves.} And, in social domains, it is important to be careful about the \textit{responsibilization} for rational response that some of these models introduce. \footnote{Differing degrees of strategic behavior, i.e. the differential breakdown of model assumptions along lines of social inequality, can itself introduce inequality, as has been studied in refinements of strategic classification \citep{hu2019disparate,Milli2019social}. \citep{miller2020strategic} points out that these models of strategic models are often structural causal models, i.e. they impose assumptions on how interventions impact covariates and future response.} 
Nonetheless, these studies highlight potential shortcomings of standard prediction formulations and the urgency of contextual assessments of prediction models embedded in the social world. 

Some recent work models the broader ecosystem of the use of automated-decision support, via composed decisions \citep{dwork2018individual}, or industry-wide use of hiring tools \citep{kleinberg2021algorithmic}.\footnote{An individual subjugated to arbitrary algorithmic decision-making in one sphere (say, child welfare) is likely also interfacing with other eligibility systems in other spheres. 
\citep{creel2022algorithmic} terms this ``the algorithmic leviathan" and  highlights how the simultaneous use of ADS across multiple sectors could lead to systemic ex/inclusion, and is of a distinct moral concern. }

\paragraph{Modeling social dynamics in causal inference }
Introducing the interventional or potential-outcomes framework introduces potentially restrictive modeling assumptions on the social world. 
One clearly vulnerable assumption is that of no-interference, that only one's own treatment affects their outcome. Interference has concrete implications for the validity of A/B testing in companies and policy conclusions from RCTs. Tackling interference typically requires some modeling of societal systems via explicit network structure, economic structure via equilibrium response or strategic response \citep{munro2021treatment,sahoo2022policy}.

While interference appears the most ``obviously" vulnerable assumption for the social world, other  assumptions are less obviously vulnerable,  such as modularity of interventions, ``well-defined interventions", and consistency. Some argue that ``well-defined interventions" are inherently conservative, in the sense of limiting interventions to local perturbations around the status quo \citep{stevenson2023cause}.\footnote{
\citet{stevenson2023cause} contends that in general, the kinds of interventions that lend themselves to high-quality program evaluations and pilots reflect the constraints of programmatic activities, rather than pursuing structural social change which might be more transformative in improving outcomes. More specifically, \citep{stevenson2023cause} considers the mixed evidence base for interventions in the criminal-legal system and finds that overall, treatment effects of published studies are usually not statistically significant, and that studies that show promise in one location struggle to scale to other locations. }  On the other hand, \textit{structural} social change can be inherently ill-suited to the framework of causal comparison, which seeks to isolate differences in context and setting to that of the intervention alone, ``holding fixed everything but the intervention". 

\subsection{Spurious correlations }
Predictive models can include unstable statistical correlations that may reflect idiosyncrasies of the data-collection environment, and do not generalize well. Causal modeling provides a precise language to describe relationships between environments and data. We describe a now-famous example.
\begin{example}[Pneumonia/asthma ``spurious correlation"]
\citep{caruana2015intelligible}. 
 studied pneumonia risk prediction. They found a candidate model  predicted asthma would \textit{reduce} the risk of pneumonia (\textit{opposite} biological mechanisms.) because of the social system generating the data: patients with asthma were admitted to a separate unit and received more intensive treatment, improving their prognosis.   
\end{example}
In this example, a purely predictive model learned in one locale can learn a so-called ``shortcut" that lacks face validity. The use of interpretable models made this error more salient, hence fixable. Though this pattern may improve in-sample predictive power, it wouldn't generalize to deployment or a new environment, and would be uninformative for improving decisions. Proposals to diagnose and remedy these (potentially with auxiliary information \citep{makar2022causally}) leverage graphical causal models. See \citet{subbaswamy2022unifying} for a graphical model of the impact of environment on the presentation of features, and graphical tools for assessing in/stability of predictors.

\section{Additional discussion on Evaluation Science}

\subsubsection{Challenges and Open Problems for RCTs of ADS in Social Settings}

While we have discussed conceptual challenges in the design of evaluations so far, there are also many practical challenges in experimental design and execution, such as low power, interference, and ethical design. 
The challenges of under-powered studies are well-documented (e.g., \cite{button2013power}); this is likely the case as effects in social settings are expected to be small. Interference also poses significant challenges. 
Interference arises naturally when interventions are scaled beyond a trial to the population at large; this is a natural concern for ADS which are often introduced for efficiency's sake for consequential decision-making at scale.\footnote{For example, a recent digital health intervention–which performed well in a pilot study–was found to be ineffective in a larger trial. This under-performance resulted from interference; as the non-profit offering the intervention had limited staff, the number of patients receiving the intervention at a point in time negatively affected the timeliness of the intervention for other patients \cite{boutilier2024randomized}.} Recent work develops experimental designs to account for interference (e.g., \cite{boyarsky2023modeling, baird2018optimal}). \citet{raji2024designing} specifically discusses interference among decision subjects induced by a common decision maker, and suggest new experimental designs that take into cognitive factors that influence human decision making. 

Qualitative analysis is just as important as quantitative analysis in evaluating interventions. Program evaluation in practice leverages both qualitative and quantitative analysis. But academic researchers in quantitative methods, with the exception of a few mixed-methods projects \citep{bergman2024creating}, rarely discuss the importance of qualitative information, which can provide invaluable insights into intervention success or failure. More directly incorporating qualitative and quantitative methods is an important future direction.

Finally, it may not be ethical or just to experiment with certain interventions. Designing experimental evaluations of ADS should always follow key ethical principles (e.g., \cite{phillips2021ethics}). Appropriate steps should be taken to protect participants from harm. For example, well-defined early stopping protocols. This is also an open area of research, as much work on experimental design at scale does not take place in consequential social settings.

\section{Additional discussion on Implementation Science}

\clearpage

\onecolumn

\bibliographystyle{abbrvnat}

\bibliography{bib/actionability-ltliu,bib/prediction-opt-decision-aware-learning,bib/predictions-and-interventions,bib/encouragement-designs,bib/model-design,bib/evaluation,bib/miscellaneous,bib/prescriptive_angles}

\end{document}